\pdfoutput=1
\documentclass[11pt]{article}
\usepackage[final]{acl}
\usepackage{times}
\usepackage{latexsym}
\usepackage[T1]{fontenc}
\usepackage[utf8]{inputenc}
\usepackage{microtype}
\usepackage{inconsolata}
\usepackage{graphicx}
\usepackage{booktabs}
\usepackage{amsmath}
\usepackage{bbm}  %
\usepackage{enumitem}  %
\title{Loop Before You Leap: What Recurrent Depth\\
Can and Cannot Buy in Data-Limited Pretraining}

\author{
  \textbf{Daichi Hotta\textsuperscript{1}} \quad
  \textbf{Yohei Kobashi\textsuperscript{2}} \quad
  \textbf{Takeshi Kojima\textsuperscript{2}} \\
  \textbf{Yusuke Iwasawa\textsuperscript{2}} \quad
  \textbf{Yutaka Matsuo\textsuperscript{2}} \\
  \textsuperscript{1}Aoyama Gakuin University \\
  \textsuperscript{2}The University of Tokyo \\
  \texttt{daichi.d.hotta@gmail.com}
}

\begin{document}
\maketitle

\begin{abstract}
Multimodal models have often scored below text-only baselines in the
BabyLM Challenge \citep{klerings2024modular,takmaz2025merging}. We test
whether additional recurrent computation helps close this gap.
Recurrence increases computation by reusing layers without adding
parameters or training data. Using a single implementation, we compare
standard and looped Transformers under a 100M-word budget, with three
seeds per main condition. Each image is represented by one feature
vector from a frozen DINOv2 encoder, used as a prefix token. Caption
words count toward the budget, so multimodal training also changes the
language data. In a pre-registered text-only comparison, a 12-layer
stack repeated four times exceeds a standard 48-layer Transformer by
5.4 BLiMP percentage points at matched inference compute. However,
multimodal training lowers BLiMP accuracy by about 3.8 points in both
architectures, with no clear mitigation from recurrence. An exploratory
loop-side no-image control, trained with the multimodal language data
and batching schedule, scores 5.7 points below the text-only baseline.
Adding the visual prefix raises accuracy by 1.85 points on average, but
this contrast is uncertain with three seeds. These results suggest that
changes in language data and its presentation are a major source of the
drop. Diagnostic tests show that
the models use visual information; its contribution to grammatical
generalization remains uncertain under this single-vector interface.
\end{abstract}

\section{Introduction}
Children acquire language from far less linguistic input than modern
language models, and they do so while perceiving the world. Motivated by
this contrast, the BabyLM Challenge limits the amount of training text
and asks how data, learning objectives, and model architectures affect
language learning under that constraint
\citep{charpentier2025findings,babylm2026cfp}.

The 2024 and 2025 BabyLM challenges included a multimodal track to explore
whether visual input could support language learning under a 100M-word
budget \citep{hu2024findings,charpentier2025findings}. Yet visual input has
not reliably improved language performance, and multimodal training can
reduce scores on grammatical benchmarks
\citep{klerings2024modular,takmaz2025merging}. This raises the question of
what prevents models from using visual information to improve language
learning.

One possibility is that integrating visual and linguistic information
requires additional sequential computation. A looped Transformer increases
this computation by repeatedly applying the same layers, without adding
parameters or training data \citep{universaltransformer,geiping2025depth}.
We therefore test whether recurrence reduces the language-performance gap
between text-only and multimodal training.

To assess how recurrence uses computation, we compare looped Transformers
with deeper standard models at matched inference compute. This comparison
asks whether, under a fixed word budget, additional computation is better
spent on layers with separate weights or on repeated use of the same layers.

In our multimodal setting, captions count toward the same word budget and
replace part of the original text corpus. This changes the mix of language
sources and genres, so the performance gap cannot be attributed to visual
input alone. We therefore use no-image controls in both architectures
to examine the contribution of the changed language data, and a
batch-matched control to separate the visual-prefix contribution from
the remaining data and batching changes in the looped model.

\begin{figure}[t]
\centering
\includegraphics[width=\linewidth]{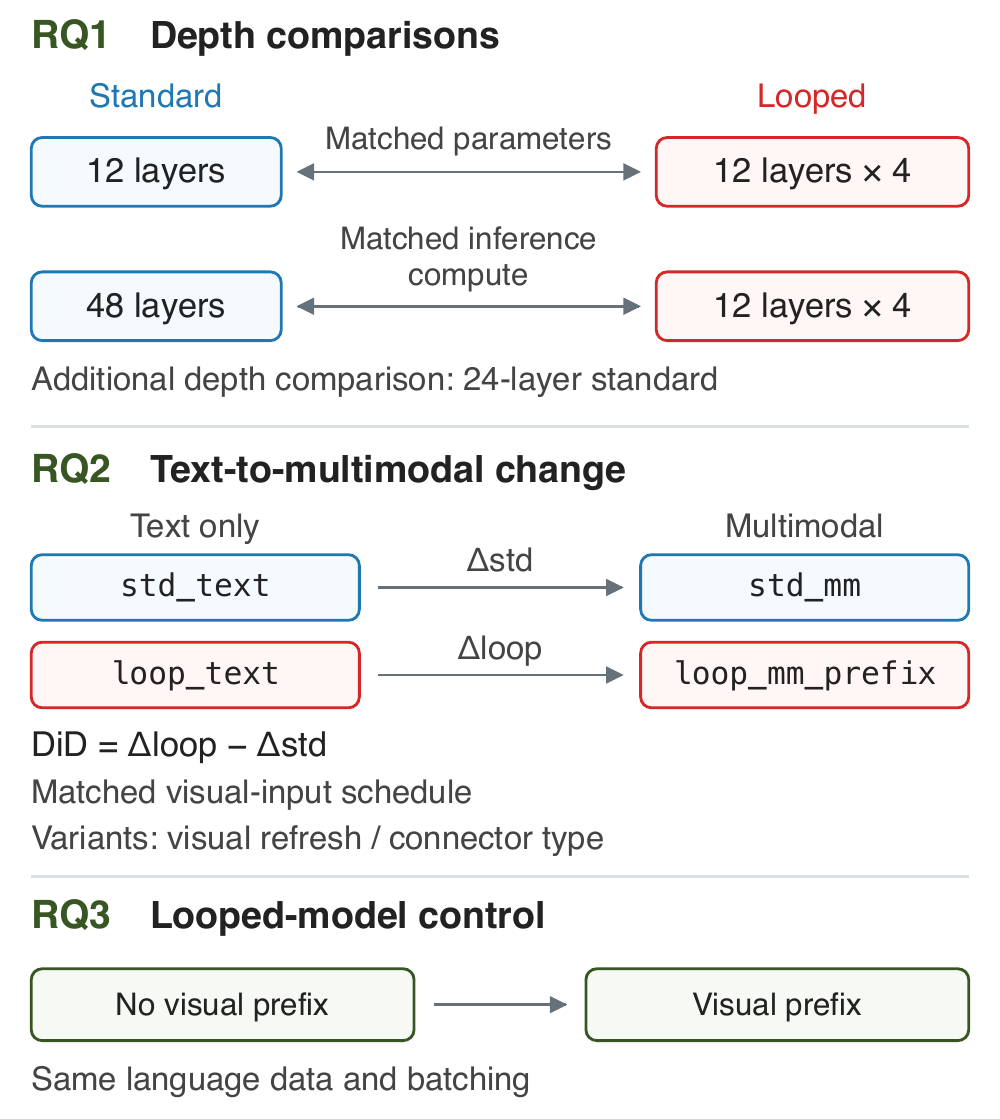}
\caption{RQ1 compares standard and looped models at matched parameter
count or inference compute. RQ2 compares their changes in accuracy from
text-only to multimodal training. RQ3 compares the looped multimodal
model with a no-image control matched in language data and batching.}
\label{fig:map}
\end{figure}

All conditions share a single implementation. The standard configuration
applies the layers once, with text-input injection and visual refresh
disabled. We compare both architectures with
and without visual input, using the same visual input scheme for the main
comparison between multimodal models. Figure~\ref{fig:map} summarizes the
three comparisons.

Each image is represented by a single feature vector from a frozen DINOv2
encoder pretrained on images alone \citep{dinov2}. The language model is
trained from scratch on text and image captions, using the visual features
to condition caption prediction.

\textbf{RQ1.} At matched inference compute, the looped model exceeds the
48-layer standard model by 5.4 BLiMP percentage points in text-only training
and 3.7 points in multimodal training. Its gains over the parameter-matched
12-layer model are much smaller, at about one point, and remain uncertain
with three seeds (Section~\ref{sec:isoflops}).

\textbf{RQ2.} Recurrence does not reliably narrow the language-performance
gap between text-only and multimodal training. The main comparison estimates
a difference of $-0.01$ BLiMP points between the performance drops, with a
95\% confidence interval of $[-1.44, 1.41]$. Tests with alternative connectors
and visual injection schedules also provide no clear evidence of mitigation
by recurrence (Section~\ref{sec:rq2}; uncertainty estimates in Limitations).

\textbf{RQ3.} Exploratory controls suggest that changes in language
data and its presentation are a major source of the BLiMP drop.
A batch-matched no-image control for the looped model scores 5.71
points below its original text-only baseline. The
visual prefix recovers 1.85 points on average, with a 95\% confidence
interval of $[-1.67,5.37]$. This exploratory decomposition combines
language-data and batching changes in its nonvisual term. The matched
decomposition is limited to the looped model, so we cannot compare
these components between architectures (Section~\ref{sec:rq3}).

\section{Related Work}
\label{sec:related}
\paragraph{Sample-efficient pretraining.}
High-performing BabyLM submissions have introduced changes to training
data, learning objectives, and model architecture
\citep{hu2024findings,charpentier2025findings}. The relationship between
training compute and performance has been less consistent: it was strong
in the 2024 analysis but varied across tracks in 2025. This motivates
controlled comparisons of how computation is allocated under a fixed
word budget.

\paragraph{Recurrent and weight-tied Transformers.}
Universal Transformers repeatedly apply a shared transformation across
depth \citep{universaltransformer}. \citet{geiping2025depth} vary the number
of repetitions of a shared block to increase computation at inference.
\citet{saunshi2025looped} find that looped language models can approach or
exceed deeper non-looped models on reasoning tasks despite worse
perplexity at matched compute. \citet{schwethelm2026isodepth} find that
replacing independent layers with shared recurrences increases validation
loss at matched training compute.

\paragraph{Recurrence meets multimodality.}
RecursiveVLM extends a pretrained multimodal model with recurrent
processing \citep{xu2026recursivevlm}. HIVE combines a pretrained recurrent
language model with visual features drawn from multiple encoder layers
and injected across iterations \citep{zhang2026hive}. We study recurrent
computation when the language model is trained from scratch under a
100M-word budget, focusing on grammatical generalization.

\paragraph{Grounding at developmental scale.}
\citet{amariucai2023multimodal} independently vary text and image quantities
to investigate whether visual input can compensate for limited language
data. \citet{takmaz2025merging} partially restore language performance by
merging multimodal models with text-only models. \citet{ganescu2025}
introduce token-wise gating to control the contribution of visual features
during language processing. Our study tests whether recurrent computation
narrows the language-performance gap between text-only and multimodal
training. Controls using identical language data then distinguish the
effect of visual input from that of changing the text corpus.

\section{Method}
\label{sec:method}

\subsection{One implementation, six conditions}
We use a single implementation of a decoder-only Transformer
\citep{transformer} for both standard and looped models. In the six main
conditions, the backbone consists of 12 layers. The standard model applies
this stack once, whereas the looped model applies it four times, sharing
weights across iterations.

The backbone follows a LLaMA-style design \citep{llama}, with pre-norm
RMSNorm, rotary position embeddings \citep{rope}, and SwiGLU feed-forward
layers. We omit bias terms and share weights between the input embedding
and output projection \citep{weighttying}.

Let $E(x)$ denote the token embeddings of text $x$, $v$ the visual feature
vector (multimodal cells only), and $P(\cdot)$ the connector that maps $v$
into model space. The forward computation applies the shared core stack $K$
times:
\begin{equation}
\label{eq:loop}
\begin{aligned}
h^{(0)} &= [\,P(v);\,E(x)\,],\\[2pt]
h^{(k)} &= \mathrm{Core}\Bigl(h^{(k-1)}\\
        &\qquad
           + \mathbbm{1}[k \le N_{\mathrm{inj}}]\,[\,P(v);\,\mathbf{0}\,]\Bigr)\\
        &\quad + a\,[\,\mathbf{0};\,E(x)\,],
\end{aligned}
\end{equation}
for $k=1,\dots,K$. The looped models add the original text embeddings
after every application of the shared core. We denote this input-injection
setting by $a$, with $a=1$ for looped models and $a=0$ for standard models.

Text-only models omit the visual prefix and visual refresh term, so
$h^{(0)}=E(x)$. With $K=1$, $a=0$, and $N_{\mathrm{inj}}=0$, the computation
reduces to a standard Transformer. The final state $h^{(K)}$ passes through
RMSNorm and the output projection to predict the next token.

\begin{figure}[t]
\centering
\includegraphics[width=\linewidth]{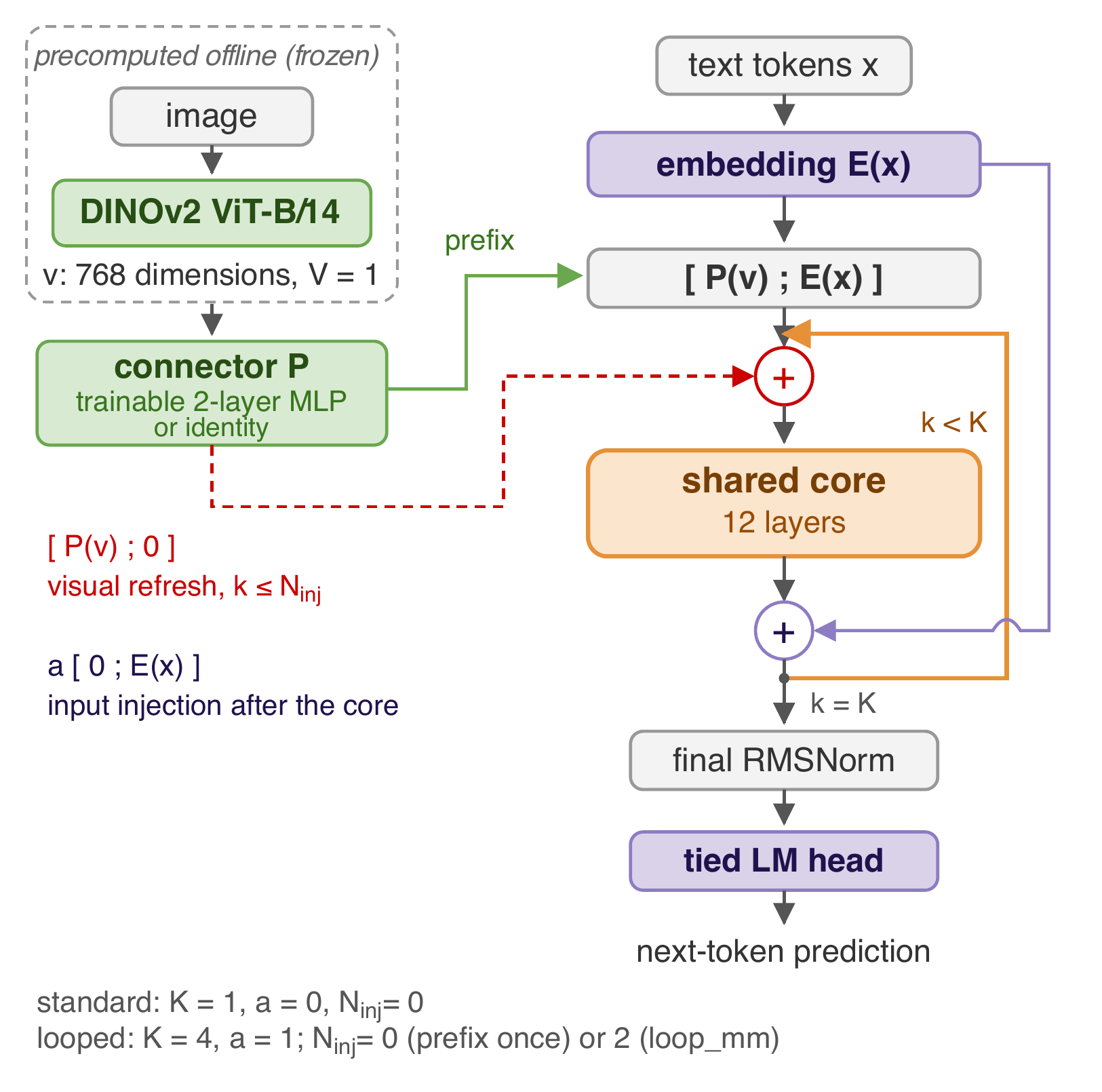}
\caption{Model architecture with a single visual prefix token from frozen
DINOv2 features. Looped models add text embeddings after every application
of the shared core. The dashed path adds the visual vector before the
first $N_{\mathrm{inj}}$ applications in \texttt{loop\_mm}.}
\label{fig:arch}
\end{figure}

\paragraph{Parameter and compute accounting.}
Repeating the shared core increases computation without adding parameters
(Table~\ref{tab:params}, Appendix~\ref{app:repro}). At $K=4$, the 12-layer
core performs 48 layer applications. We compare this model with a 12-layer
standard Transformer at matched parameter count and a 48-layer standard
Transformer at matched inference compute.

\subsection{Visual grounding by offline, frozen features}
\label{sec:vision}

Each image is represented by a single 768-dimensional feature vector from
a frozen DINOv2 ViT-B/14 encoder \citep{dinov2}. We use the precomputed
vectors supplied with the BabyLM multimodal data. The encoder was
pretrained on images only; raw images are not processed during
language-model training.
Figure~\ref{fig:arch} shows the architecture.

A connector $P$ maps the visual vector into the language model's embedding
space. For image--caption examples, we prepend the resulting vector to the
caption embeddings as a single visual token. The model is trained to
predict text and caption tokens; text-only examples have no visual prefix.

By default, $P$ is a two-layer MLP with a GELU activation between two
$768 \times 768$ linear transformations \citep{llava,llava15}. We also test
an identity connector that passes the visual vector through unchanged,
since its dimension already matches the language model's embedding size.

With $N_{\mathrm{inj}}=0$, the visual vector is supplied only as the initial
prefix. With $N_{\mathrm{inj}}>0$, the same projected vector is also added
at the visual-prefix position before each of the first $N_{\mathrm{inj}}$
applications of the shared core (Eq.~\ref{eq:loop}).

\section{Experimental Setup}
\label{sec:setup}

\subsection{Conditions and seeds}
\label{sec:conditions}
The four core conditions combine standard or looped models with text-only
or multimodal training (Figure~\ref{fig:map}). The looped models use $K=4$.
In multimodal training, \texttt{std\_mm} supplies the visual vector only as
an initial prefix, whereas \texttt{loop\_mm} also adds it at the prefix
position before the first two iterations. We fixed $N_{\mathrm{inj}}=2$
a priori; it was not tuned.

Two additional conditions test the visual injection schedule and connector.
In \texttt{loop\_mm\_prefix}, the visual vector is supplied only as an initial
prefix, matching \texttt{std\_mm}. In \texttt{std\_mm\_id},\footnote{Named
\texttt{std\_mm\_naive} in the released configurations.} an identity
connector replaces the MLP, passing the visual vector through unchanged.
Each condition is one Hydra configuration, reproduced with a single argument.

We train each of these six conditions with three seeds (42, 43, and 44),
stopping when cumulative exposure reaches 1B words. The peak learning rate
is selected on \texttt{std\_text} and reused across these six conditions.

\subsection{Data and word budget}
\label{sec:data}
The text-only conditions use 100M words from the BabyLM 2026 Strict corpus
\citep{babylm2026cfp}. The multimodal conditions use 50M words from the
same corpus and 50M caption words, keeping the total language budget at
100M words.\footnote{Exact whitespace-delimited counts are 100,000,000
words for text-only training, and 49,999,988 text words plus 49,999,996
caption words for multimodal training.}
Each model is trained until cumulative exposure reaches 1B words,
counting repeated presentations of the training data. For multimodal
models, this total includes words from both text and caption batches.

The captions and precomputed visual features come from the 2024 BabyLM
multimodal release (OSF project \texttt{ad7qg}). Our caption set contains
3,037,190 examples from Localized Narratives \citep{localizednarratives}
and Conceptual Captions 3M \citep{cc3m}.

For the multimodal conditions, we take the first 50M words of the text
corpus in its stored order. This changes the proportions of language
sources in the retained text. Together with replacing the remaining text
with captions, this produces what we call the \emph{composition shift}.

CHILDES \citep{childes2000} accounts for 56.8\% of the retained text,
compared with 28.4\% of the full text corpus. The remaining words come from
BNC spoken dialogue (15.2\%) and Gutenberg (27.9\%). OpenSubtitles, Simple
Wikipedia, and Switchboard are absent from the retained text. These
proportions are measured by word count against the official domain files
\citep{babylm2026cfp}.

We train a byte-level BPE tokenizer \citep{bpe,gpt2} with a vocabulary of
16,000 tokens on the text corpus. The same tokenizer is used across all
conditions without further training. We release the assembled corpus as a
Hugging Face dataset with a datasheet \citep{gebru2021datasheets} recording
provenance, domain composition, and per-file manifests
(Appendix~\ref{app:repro}).

\subsection{Optimization: a words-seen--driven schedule}
\label{sec:schedule}
We use AdamW \citep{adamw} with $\beta_1=0.9$, $\beta_2=0.95$, and weight
decay of 0.1. Training uses bfloat16 precision, and the gradient norm is
clipped at 1.0. For the six main conditions, we use micro-batches of 64
sequences, with text sequences of up to 512 tokens. Gradients are
accumulated over eight micro-batches before each optimizer update.

The peak learning rate is $1.5\times10^{-3}$, selected from five candidates
evaluated in 100M-word training runs on \texttt{std\_text}.\footnote{The
candidates span $3\times10^{-4}$ to $2\times10^{-3}$.}
We schedule the learning rate by cumulative words seen. It increases
linearly over the first 15M words, then follows a cosine decay
\citep{sgdr} to 10\% of its peak at 1B words. For the decay phase,
\begin{equation}
\label{eq:lr}
\eta(w)=\eta_{\max}\Bigl[\rho+\tfrac{1-\rho}{2}
        \bigl(1+\cos\pi\,p(w)\bigr)\Bigr],
\end{equation}
where $w$ is cumulative words seen,
$p(w)=\frac{w-w_{\text{warm}}}{w_{\max}-w_{\text{warm}}}$,
$w_{\text{warm}}=15$M, $w_{\max}=1$B, and $\rho=0.1$.

Captions are kept as separate sequences, whereas text-only training packs
text into fixed-length blocks. Multimodal training therefore requires more
optimizer updates to reach the same word exposure.\footnote{At seed 42,
\texttt{std\_mm} takes 13,895 updates and \texttt{std\_text} takes 6,827.}
Scheduling by words seen aligns the learning rate across conditions at the
same cumulative exposure.

\subsection{Checkpoints and evaluation}
\label{sec:eval}
We save checkpoints at 28 cumulative-word milestones between 1M and 1B
words. Formats and loading details are in Appendix~\ref{app:repro}.
We evaluate the models using a fixed revision of the official BabyLM
pipeline with causal language-model scoring.

We report BLiMP \citep{blimp} accuracy on the full filtered evaluation set,
with a breakdown by linguistic category (\texttt{linguistics\_term}).
For the six main conditions, we report the mean and standard deviation
across three seeds.\footnote{The public leaderboard hosts single-seed
submissions of the four core cells (seed 42, except seed 43 for
\texttt{loop\_mm}); the extension cells are analysis-only and not submitted.
This paper reports three-seed means; the board's own aggregation can differ
in the last digit (76.62 vs.\ our 76.61 for \texttt{loop\_text}).}
Our aggregation reproduces the official average (75.23 for
\texttt{std\_text}, seed 42) as a correctness check.

\subsection{Analysis: the interaction as a difference-in-differences}
\label{sec:analysis}
For RQ2, we compare the change in BLiMP accuracy from text-only to multimodal
training between standard and looped models. Our primary comparison uses
\texttt{loop\_mm\_prefix}, so that both multimodal models receive visual
input only as an initial prefix. We compute this contrast as a
difference-in-differences, with condition names denoting BLiMP accuracy:
\begin{equation}
\label{eq:did}
\begin{aligned}
\Delta_{\mathrm{DiD}}
={}& (\texttt{loop\_mm\_prefix}-\texttt{loop\_text})\\
 {}-{}& (\texttt{std\_mm}-\texttt{std\_text}).
\end{aligned}
\end{equation}
Positive values indicate a smaller text-to-multimodal drop for the looped
model; negative values indicate a larger drop.

We also compute the contrast with \texttt{loop\_mm} in place of
\texttt{loop\_mm\_prefix} to assess sensitivity to visual refresh. A second
set of contrasts uses \texttt{std\_mm\_id} as the standard multimodal baseline.

Switching from text-only to multimodal training changes both the visual
input and the language data. A corpus effect that is equal in both
architectures cancels in the difference-in-differences. Differences in how
the architectures respond to the changed corpus remain part of the
contrast. The architectural contrast includes both recurrence and
text-input injection, whose individual contributions are not separated in
this design.

\section{Results}
\label{sec:results}

\begin{table}[t]
\centering
\small
\begin{tabular}{lcc}
\toprule
 & text & multimodal \\
\midrule
standard ($K{=}1$) & $75.21{\pm}0.68$  & $71.37{\pm}0.29$ \\
looped ($K{=}4$, re-inj.) & $76.16{\pm}0.45$  & $72.38{\pm}0.83$ \\
\midrule
looped, prefix-once & --- & $72.31{\pm}0.63$ \\
identity connector ($K{=}1$) & --- & $72.48{\pm}0.69$ \\
identity connector ($K{=}4$) & --- & $72.14{\pm}0.15$ \\
\midrule
std2x (24 layers)   & $73.84{\pm}0.25$ & $70.96{\pm}0.88$ \\
std4x (48 layers)   & $70.80{\pm}0.60$ & $68.64{\pm}0.29$ \\
\bottomrule
\end{tabular}
\caption{BLiMP accuracy (\%) on the full evaluation set, reported as
mean $\pm$ standard deviation across three seeds. All multimodal
conditions use an initial visual prefix; ``re-inj.''\ also adds the visual
vector before the first two core applications.}
\label{tab:results}
\end{table}

\begin{table}[t]
\centering
\small
\setlength{\tabcolsep}{3pt}
\begin{tabular}{lcc}
\toprule
 & vs.\ \texttt{std\_mm} & vs.\ \texttt{std\_mm\_id} \\
\midrule
penalty (standard)            & $-3.84$ & $-2.74$ \\
penalty (looped, re-inj.)     & $-3.79$ & $-3.79$ \\
penalty (looped, prefix-once) & $-3.86$ & $-3.86$ \\
\midrule
$\Delta_{\mathrm{DiD}}$ (re-injected)   & $+0.06$ & $-1.05$ \\
$\Delta_{\mathrm{DiD}}$ (\textbf{matched}) & $-0.01$ & $-1.12$ \\
\bottomrule
\end{tabular}
\caption{Changes in BLiMP accuracy from text-only to multimodal training
and their interactions, in percentage points (three-seed means). Positive
$\Delta_{\mathrm{DiD}}$ indicates a smaller drop for the looped model.
The primary contrast is the ``matched'' row under \texttt{std\_mm}, where
both multimodal models use only an initial visual prefix
(Eq.~\ref{eq:did}).}
\label{tab:did}
\end{table}

\subsection{RQ1: iteration beats added parameter depth}
\label{sec:isoflops}

\paragraph{A parameter-matched, modality-independent depth gain.}
At matched parameter count, the looped text-only model scores 76.16 on
BLiMP, compared with 75.21 for the 12-layer standard Transformer. The
0.95-point difference is not statistically significant with three seeds
per condition (two-sided Welch's test, $p\approx0.12$).\footnote{Welch's
$t\approx2.0$; Appendix~\ref{app:seeds} reports per-seed values.}
In multimodal training, \texttt{loop\_mm\_prefix} exceeds \texttt{std\_mm}
by 0.94 BLiMP points on average, similar to the text-only difference.
The difference is positive in all three seed pairings: 0.72, 0.61, and
1.48 points.

The visual-refresh variant, \texttt{loop\_mm}, exceeds \texttt{std\_mm} by
1.01 BLiMP points, although this comparison also changes the visual
injection schedule. Against the standard model with an identity connector,
\texttt{std\_mm\_id}, \texttt{loop\_mm} scores 0.10 points lower.

\paragraph{Iso-FLOPs baselines: looping beats added depth.}
We compare the looped model with standard Transformers containing 24 and
48 independent layers. The 48-layer model matches the looped model's
inference compute. We registered the comparison and its statistical
decision rule on the Open Science Framework before observing the final
results from these deeper standard models.\footnote{Study-plan registration,
accessible through the view-only link:
\url{https://osf.io/438b9/?view_only=44cd0d22cb19498a9282ed806de06c02}.}

The 24- and 48-layer models have approximately 180M and 350M parameters,
respectively. We train each depth in both text-only and multimodal
conditions with three seeds. A separate peak learning rate is selected for
each depth through 100M-word training probes.\footnote{For the 48-layer
models, we use micro-batches of 32 sequences and accumulate gradients over
16 micro-batches, preserving the nominal batch size of 512 sequences per
update.}

At matched inference compute, the looped model outperforms the
48-layer standard model by 5.36 BLiMP points in text-only training and
3.67 points in multimodal training. Across the standard text-only models,
BLiMP falls monotonically as depth is added ($75.21 \to 73.84 \to 70.80$;
Table~\ref{tab:results}), while the looped model reaches $76.16$ by
iterating 97M parameters.

Deeper standard models have lower training loss but also lower BLiMP
accuracy. For seed 42, average training loss is 1.88 with 12 layers, 1.63
with 24 layers, and 1.26 with 48 layers.\footnote{Averaged over the final
eight logged values.}

Our preregistered test asks whether multimodal training increases the
looped model's advantage over the 48-layer standard model. We subtract the
text-only advantage from the multimodal advantage:
\[
\begin{aligned}
\mathrm{DiD}_{\mathrm{iso}}
={}& (\texttt{loop\_mm\_prefix}-\texttt{std4x\_mm})\\
 {}-{}& (\texttt{loop\_text}-\texttt{std4x\_text}).
\end{aligned}
\]
We pair runs with the same seed. The mean contrast is $-1.69$ BLiMP points,
with a 90\% confidence interval of $[-3.22,-0.16]$.\footnote{The interval
uses a $t$ distribution with df$=2$, following the registered procedure;
the registered decision margin is $\delta=1.8$, based on the paired
95\% confidence interval of the main-grid interaction. The per-seed
contrasts are $-1.91$, $-2.47$, and $-0.69$.}
This meets the preregistered \emph{Reversed} criterion: the looped model's
advantage is smaller under multimodal training. The 95\% confidence interval
is $[-3.95,0.57]$ and includes zero.\footnote{On the log-odds scale, the
90\% confidence interval is also negative ($-0.17$ to $-0.02$).}

Deeper standard models have lower text-only BLiMP scores and smaller drops
under multimodal training (Table~\ref{tab:results}). Their smaller drops
may partly reflect their lower text-only performance. This is a post-hoc
explanation that our design cannot distinguish from a cost specific to the
looped architecture.

The corresponding contrast against the 24-layer standard model is $-0.97$
BLiMP points. Appendix~\ref{app:ksweep} reports additional experiments
varying $K$, with one seed each for the $K=2$ and $K=8$ runs.

\subsection{RQ2: recurrence does not clearly reduce the multimodal drop}
\label{sec:rq2}

\paragraph{Similar drops across architectures.}
Compared with text-only training, BLiMP accuracy drops by 3.84 points for
\texttt{std\_mm} and 3.86 points for \texttt{loop\_mm\_prefix}. The drop for
\texttt{loop\_mm}, which refreshes the visual prefix, is 3.79 points
(Tables~\ref{tab:results} and~\ref{tab:did}).

The primary difference-in-differences estimate is $-0.01$ BLiMP points,
with a 95\% confidence interval of $[-1.44,1.41]$.\footnote{This interval
pools the error across twelve runs. The seed-paired interval is
$[-1.83,1.80]$; see Limitations for both estimates.}
We find no clear evidence that the looped model reduces the drop
associated with multimodal training.

Using \texttt{loop\_mm} instead of \texttt{loop\_mm\_prefix} gives a
difference-in-differences estimate of $+0.06$ points. When
\texttt{std\_mm\_id} is used as the standard multimodal baseline, the
estimates are $-1.05$ points with \texttt{loop\_mm} and $-1.12$ points
with \texttt{loop\_mm\_prefix}.

Across the 13 linguistic categories, the mean primary contrasts range from
$-1.66$ to $1.48$ BLiMP points. The sign changes across seeds in 10 of the
13 categories (Figure~\ref{fig:did}, Appendix~\ref{app:phenom}).

\paragraph{Connector comparison.}
In the standard model, the identity connector scores 72.48 on BLiMP,
compared with 71.37 for the learned MLP. The 1.11-point difference is not
statistically significant with three seeds per condition (two-sided
Welch's test, $p\approx0.09$).

We repeat this comparison in the looped model with three seeds, replacing
only the MLP in \texttt{loop\_mm\_prefix} with an identity connector. This
condition, \texttt{loop\_mm\_naive}, scores 72.14 on BLiMP, compared with
72.31 for \texttt{loop\_mm\_prefix}. These comparisons provide no clear
evidence that the learned MLP improves BLiMP accuracy in either architecture.

\paragraph{Visual-refresh comparison.}
With $K=4$ and the training data held fixed, visual refresh changes BLiMP
accuracy by $+0.07$ points on average. The per-seed differences are $-0.78$,
$+1.55$, and $-0.56$ points, providing no clear evidence of a benefit from
visual refresh.

\subsection{RQ3: what the penalty is made of}
\label{sec:rq3}

\paragraph{A matched control for the looped model.}
We train a no-image control for the looped model, keeping the language
data, batches, word schedule, and initialization of
\texttt{loop\_mm\_prefix} and omitting only the visual prefix. Its mean
BLiMP accuracy is 70.46 across three seeds.\footnote{The per-seed scores
are 70.12, 71.38, and 69.88. All three conditions in this decomposition
were evaluated at the same pipeline revision
(Appendix~\ref{app:packedcontrols}).}
Adding the visual prefix raises BLiMP by 1.85 points on average, with
positive differences in all three seeds (2.23, 0.28, and 3.04) and a
95\% confidence interval of $[-1.67,5.37]$.

The original looped text-only model scores 5.71 points above the
no-image control, with a 95\% confidence interval of $[2.73,8.68]$.
This difference includes changes in language data, batch construction,
and the number of optimizer updates. The means suggest that the visual
prefix partly offsets the cost of this changed training recipe in the
looped model, but its contribution remains uncertain with three seeds.
Removing the prefix also removes its sequence position, so the contrast
does not isolate image content from prefix mechanics.

\paragraph{Other controls and their scope.}
We also train no-image controls on the multimodal conditions' language
data using the text-only procedure of packing tokens into fixed-length
blocks. The standard model scores 70.56, compared with 75.21 on its
original text corpus; the looped model scores 71.31, compared with
76.16 (Table~\ref{tab:results}; Appendix~\ref{app:packedcontrols}). These results support a
contribution from the changed language data even without visual input.
Packing uses the same procedure, but the number of updates can change
under the fixed word budget. Comparisons with multimodal training also
change the batching procedure and therefore do not isolate visual input.

Together, the controls suggest that changes in language data and its
presentation are a major source of the drop. The matched loop-side
control provides the direct decomposition; the packed controls provide
supporting evidence in both architectures.

The batch-matched no-image decomposition is limited to the looped
model. We therefore cannot compare the visual-prefix and
data-and-batching contributions between architectures.

\paragraph{The penalty emerges over training.}
The multimodal models briefly lead early in training, but the text-only
models overtake them as training continues. These fast-eval curves cover
28 checkpoints and use one submitted seed per condition, so we treat them
as descriptive (Appendix~\ref{app:traj}).

\paragraph{The grounded models do use vision: a shuffled-features
 diagnostic.}
To test whether the trained models use visual information, we score the
same 50,000 captions with correct, shuffled, and zeroed image features.
This changes the visual input at evaluation time while keeping the model
and captions fixed (Appendix~\ref{app:shufflediag}).

Shuffling the image features increases caption prediction loss (negative
log-likelihood, NLL) by 1.45 nats per token for \texttt{std\_mm}.
The increases are 1.52 for \texttt{loop\_mm} and 1.49 for
\texttt{loop\_mm\_prefix}.\footnote{Cohen's $d\approx1.9$ for
\texttt{std\_mm}; the standard deviation of the NLL increase across seeds
is at most 0.01 for the looped variants.}
For every evaluated model and seed, shuffling increases the loss for at
least 98.6\% of captions. This shows that the models use visual information
when predicting captions.

We also evaluate visual question answering (VQA) with correct, shuffled,
and zeroed image features. We follow the multimodal track's evaluation
protocol but extract the image features ourselves, so these results are
diagnostic rather than leaderboard scores (Appendix~\ref{app:vqa}).
All four multimodal conditions score above the 12.5\% chance level.
Correct image features yield higher accuracy than shuffled features in
all twelve model--seed combinations, with gains of 2.4 to 5.6
points.\footnote{McNemar $p<10^{-25}$ in each comparison.}
No looped condition exceeds \texttt{std\_mm}'s mean VQA accuracy of 47.4\%.
With standard deviations across seeds reaching 4.8 points, these results
provide no clear evidence of a VQA benefit from looping.

\begin{table}[t]
\centering
\small
\setlength{\tabcolsep}{4pt}
\begin{tabular}{lrrrr}
\toprule
Task & pen.\ (std) & pen.\ (loop) & DiD & DiD$_{\text{m}}$ \\
\midrule
BLiMP            & $-3.84$ & $-3.79$ & $+0.06$ & $-0.01$ \\
BLiMP Suppl.     & $-3.66$ & $-4.95$ & $-1.29$ & $-0.08$ \\
EWoK             & $+0.04$ & $-0.93$ & $-0.96$ & $-0.64$ \\
COMPS            & $-0.62$ & $-0.01$ & $+0.61$ & $-0.10$ \\
Entity tracking$^\dagger$ & $-0.42$ & $+1.85$ & $+2.28$ & $+1.88$ \\
\bottomrule
\end{tabular}
\caption{Changes in zero-shot accuracy from text-only to multimodal
training, in percentage points (three-seed means). The pen.\ (loop) and
DiD columns use \texttt{loop\_mm}, whereas DiD$_{\mathrm{m}}$ uses
\texttt{loop\_mm\_prefix} with the same visual-input schedule as
\texttt{std\_mm}. Both interactions use \texttt{std\_mm} as the multimodal
standard baseline, and positive values indicate a more favorable change
for the looped model.
$^\dagger$All conditions score below the 20\% chance level on Entity tracking.}
\label{tab:pertask}
\end{table}

\paragraph{Beyond BLiMP: different effects across tasks.}
Multimodal training lowers accuracy on BLiMP Supplement
\citep{warstadt2023findings} as it does on BLiMP
(Table~\ref{tab:pertask}). The changes are smaller on EWoK, which tests
world knowledge \citep{ewok2024}, and COMPS, which tests conceptual
knowledge \citep{comps2023}. These small changes do not establish that
these abilities are unaffected by multimodal training.

On BLiMP Supplement and EWoK, the mean interactions are negative with
both matched visual input and visual refresh (Table~\ref{tab:pertask}).
With visual refresh, the EWoK interaction is negative in all three seeds,
whereas BLiMP Supplement shows greater variation across seeds.\footnote{Across
seeds, the visual-refresh contrasts have standard deviations of 2.09 points
for BLiMP Supplement and 0.67 for EWoK.}

Entity tracking \citep{entitytracking} yields positive mean interactions,
but all conditions score below the 20\% chance level. We therefore do not
interpret these interactions as evidence that looping mitigates the
multimodal performance drop.

Reading, age of acquisition (AoA), and GLUE were evaluated only for the
submitted checkpoints and are not analyzed here.
Appendix~\ref{app:leaderboard} provides comparisons with the official
leaderboard baselines for context.

\section{Discussion}
\label{sec:discussion}

\paragraph{Recurrence and the multimodal gap.}
Compared with the 12-layer standard model, the looped model shows a similar
BLiMP gain in text-only and multimodal training. It does not clearly narrow
the gap between the two training conditions.

Multimodal training replaces half of the original text with captions,
changes the source distribution of the retained text, and changes how
batches are constructed (Section~\ref{sec:data}). These changes may
reduce exposure to some of the linguistic patterns tested by BLiMP.
The loop-side matched control quantifies their combined contribution:
the no-image condition scores 5.71 points below the original text-only
model, and the visual prefix recovers 1.85 points on average. The latter
contrast remains uncertain. Together with the packed no-image controls,
these results suggest that changes in language data and its presentation
are a major source of the drop. The missing standard-side control
prevents an attribution of the overall interaction to differences in
these components (Section~\ref{sec:rq3}).

The models use visual information to predict captions and answer
questions about images. This does not by itself establish an improvement
in grammatical generalization. In the looped model, the improvement over
its matched no-image control is positive in every seed but remains
uncertain.

\paragraph{The scope of the depth gain.}
The gain over the 12-layer standard model is modest. In our main multimodal
comparisons, \texttt{loop\_mm} scores 72.38 and remains below the
identity-connector standard model, \texttt{std\_mm\_id}, at 72.48.

The comparison with the 48-layer standard model suggests that how
computation is allocated matters under this data budget
(Section~\ref{sec:isoflops}). Weight sharing may help limit overfitting by
allowing additional computation without adding parameters.
On 20,000 held-out captions (Appendix~\ref{app:traj}), the looped text-only
model has an NLL of 5.83 nats per token. The 24- and 48-layer text-only
standard models have NLLs of 6.35 and 7.24, respectively.

\paragraph{The connector.}
Removing the learned MLP does not restore text-only BLiMP performance in
either architecture. This comparison changes the visual connector and
does not isolate the contribution of visual input. The experiments with
richer visual inputs in Appendix~\ref{app:exploratory} show no clear
BLiMP benefit. Each condition uses one seed, so these results are
preliminary.

\paragraph{From compute to mechanism.}
Gradually shifting from caption-heavy to text-heavy training yields 72.70
on BLiMP, compared with 72.31 for a constant mixture. Reversing this
schedule yields 66.62. Each direction is evaluated with three seeds
(Appendix~\ref{app:curriculum}). This asymmetry suggests that the language
distribution encountered late in training matters. Ending with
caption-heavy training lowers BLiMP accuracy, while ending with text-heavy
training offers no clear improvement over a constant mixture.

Our visual input is a single global image vector paired with a caption.
It does not explicitly represent how an utterance relates to objects and
events over time. \citet{pinker1984language} proposes that semantic cues can
help learners identify grammatical categories. \citet{gleitman1990structural}
argues that sentence structure can help learners infer verb meanings.

We hypothesize that grounding helps grammatical generalization only when
the model can connect parts of an utterance to the objects and events they
describe. Testing this hypothesis requires explicit mechanisms for learning
these connections.

\section{Conclusion}
We studied recurrent Transformers under a 100M-word training-data budget.
At matched inference compute, the looped model outperformed the 48-layer
standard Transformer on BLiMP in both text-only and multimodal training.
With a single global image vector as visual input, we found no clear
evidence that looping reduces the text-to-multimodal BLiMP drop. The
primary difference-in-differences estimate was $-0.01$ points, with a 95\%
confidence interval of $[-1.44,1.41]$.

Exploratory controls suggest that changes in language data and its
presentation are a major source of the BLiMP drop. For the looped model,
a matched no-image control separates
the change in language data and batching from the visual-prefix
contribution. The no-image control scores 5.71 points below the original
text-only baseline, while adding the visual prefix recovers 1.85 points
on average, with substantial uncertainty across three seeds. A matched
standard-model control is needed to compare these components
between architectures.
Future work should test whether explicitly linking linguistic structure to
objects and events in visual input improves grammatical generalization.

\section*{Limitations}

\paragraph{Statistical power.}
Each main condition was trained with three random seeds. This limits our
ability to detect small differences between architectures. In particular,
a primary interaction estimate near zero does not establish that looping
has no effect on the multimodal BLiMP drop.
The pooled 95\% confidence interval for the primary interaction is
$[-1.44,1.41]$ BLiMP points.\footnote{Pooling the error across the twelve
runs gives the interval reported above. Computing the interaction
separately for each of the three seeds and using a paired analysis gives
a wider 95\% confidence interval of $[-1.83,1.80]$.}
This interval allows both a reduction and an increase in the multimodal
drop \citep{lakens2017equivalence}. Its upper end corresponds to a reduction
of about 37\% in the 3.84-point drop observed in the standard model.
We report several secondary comparisons without adjusting for multiple
testing, including connector variants and breakdowns by linguistic
phenomenon and task. These analyses are exploratory, and differences
close to the significance threshold should not be treated as confirmed
effects.

\paragraph{Language data and visual input.}
Multimodal training replaces half of the text-only corpus with captions
and changes the source distribution of the retained text
(Section~\ref{sec:data}). The resulting BLiMP drop therefore combines
changes in language data with the addition of visual input. Our primary
interaction measures how the two architectures respond differently to
this combined change.
The effect of the language-data change cancels in this contrast only if
it changes BLiMP accuracy by the same amount in both architectures.
The batch-matched no-image control for the looped model
(Section~\ref{sec:rq3}) separates the visual prefix from the rest of the
multimodal recipe within that model. The remaining term combines
language-data and batching changes, because text-only training packs
text into blocks and uses fewer optimizer updates. This matched
comparison is limited to the looped model, so we cannot determine
whether either component differs between architectures.
Appendix~\ref{app:packedcontrols} reports the loop-side per-seed values,
together with packed controls that use a different batching procedure.
The looped model also adds the original text embeddings after each
application of the shared layers. Our comparison therefore does not
separate the contribution of repeated layer applications from that of
text-input injection.

\paragraph{Interpreting BLiMP scores.}
We assess grammatical generalization using BLiMP and its Supplement.
Lower scores may reflect weaker grammatical generalization, a mismatch
between the language encountered during training and the benchmark
sentences, or both. Our experiments do not fully distinguish these
explanations.
Future evaluations should test grammatical contrasts in a wider range of
language, including conversation and captions. This would help assess
whether the observed drop extends beyond the kinds of sentences used in
BLiMP.

\paragraph{Visual evaluation.}
Our visual question answering evaluation is diagnostic
(Appendix~\ref{app:vqa}). It covers one task with three seeds per condition
and substantial variation across seeds. We also use features extracted
ourselves rather than the officially distributed features, so these
scores are not directly comparable to leaderboard results.

\paragraph{Visual representation.}
Our main experiments represent each image with a single global feature
vector. The results may differ with visual inputs that preserve more
spatial detail, such as patch-level features
\citep{ganescu2025,charpentier2025findings}. Our exploratory experiments
with richer visual inputs use only one seed per condition and do not
resolve this question (Appendix~\ref{app:exploratory}).

\paragraph{The identity connector requires matched widths.}
The identity connector is possible because the visual features and the
language model's hidden states both have 768 dimensions. When these
dimensions differ, a mapping between them is required, so our findings do
not imply that a learned projection can generally be omitted.

\paragraph{Model and training settings.}
We use a single decoder-only model family, tokenizer, and training
objective. The main looped models use four iterations, and experiments
with two or eight iterations use only one seed per condition
(Appendix~\ref{app:ksweep}). The findings may therefore change with other
model designs, loop counts, or data budgets.
The main looped models use the learning rate selected for the 12-layer
standard model. We selected learning rates separately for the deeper
standard models, but kept weight decay, dropout, and warmup fixed across
depths. Further tuning for each architecture could change the comparison.
Our learning-rate selection uses training loss from short probe runs.
A setting that minimizes this loss need not give the best BLiMP accuracy
at the end of training.
At 100M words, both the 48-layer standard model and the looped model
score higher on BLiMP at $10^{-3}$ than at their main-run learning
rates.\footnote{For seed 42, the 48-layer model scores 66.02 at $10^{-3}$
and 64.03 at its main-run rate of $6\times10^{-4}$. The looped model
scores 63.65 at $10^{-3}$ and 61.20 at its shared main-run rate of
$1.5\times10^{-3}$. At this probe scale, the gains are similar in size
for both models. These probes do not establish the effect of further
learning-rate tuning at 1B words.}
All models were trained without dropout. Stronger regularization might
improve the deeper standard models, but we did not test this possibility.

\paragraph{Training-time feature controls.}
We attempted a training-time control with the visual features set to zero
while retaining the prefix token. All three seeds produced NaN losses
early in training, leaving no usable comparison. An earlier version of
this paper also reported a control that shuffled visual features across
captions. We no longer report it because we could not verify that the
shuffle was applied during training.

\paragraph{Proxy multimodal data.}
We use text from the 2026 BabyLM Strict corpus, but image--caption data
and visual features from the official 2024 multimodal release.
A different caption corpus could change both the multimodal BLiMP drop
and the interaction between architecture and training modality.

\section*{Ethics Statement}
We use the 2026 BabyLM Strict text corpus and the officially distributed
image--caption data from the 2024 multimodal track. The captions originate
from Conceptual Captions and Localized Narratives.
The 2026 Strict text corpus was filtered to reduce toxic content
\citep{babylm2026cfp}. Nevertheless, our training data may still contain
harmful content or social biases, which the models may reproduce.
Our models are intended for research on language learning from limited
data. They are not intended for deployment in user-facing applications.

For the exploratory experiments, we retrieved images associated with the
original caption datasets and extracted richer visual features
(Appendix~\ref{app:exploratory}). Some images were no longer available, so
these experiments used a reduced subset.
Our released training corpus contains text, captions, and provenance
records, but no raw images. We direct users to the official BabyLM release
for the precomputed visual features used in the main experiments.

This study uses existing datasets and does not involve experiments with
human participants. Personal information may still be present in the
source corpora.

\section*{Acknowledgments}
We thank the anonymous reviewers for their helpful comments.

\bibliography{custom}

\begin{thebibliography}{39}
\providecommand{\natexlab}[1]{#1}

\bibitem[{Amariucai and Warstadt(2023)}]{amariucai2023multimodal}
Theodor Amariucai and Alex Warstadt. 2023.
\newblock Acquiring linguistic knowledge from multimodal input.
\newblock In \emph{Proceedings of the BabyLM Challenge at the 27th Conference
  on Computational Natural Language Learning (CoNLL)}.

\bibitem[{Charpentier et~al.(2025)Charpentier, Choshen, Cotterell, Gul, Hu,
  Liu, Jumelet, Linzen, Mueller, Ross, Shah, Warstadt, Wilcox, and
  Williams}]{charpentier2025findings}
Lucas Georges~Gabriel Charpentier, Leshem Choshen, Ryan Cotterell, Mustafa~Omer
  Gul, Michael~Y. Hu, Jing Liu, Jaap Jumelet, Tal Linzen, Aaron Mueller,
  Candace Ross, Raj~Sanjay Shah, Alex Warstadt, Ethan~Gotlieb Wilcox, and Adina
  Williams. 2025.
\newblock \href {https://doi.org/10.18653/v1/2025.babylm-main.28} {Findings of
  the third {BabyLM} challenge: Accelerating language modeling research with
  cognitively plausible data}.
\newblock In \emph{Proceedings of the First BabyLM Workshop}, pages 399--420.
  Association for Computational Linguistics.

\bibitem[{Charpentier and Samuel(2024)}]{gptbert}
Lucas Georges~Gabriel Charpentier and David Samuel. 2024.
\newblock \href {https://aclanthology.org/2024.conll-babylm.24/} {{GPT} or
  {BERT}: why not both?}
\newblock In \emph{The 2nd BabyLM Challenge at the 28th Conference on
  Computational Natural Language Learning (CoNLL)}, pages 262--283, Miami, FL,
  USA. Association for Computational Linguistics.

\bibitem[{Choshen et~al.(2026)Choshen, Cotterell, Gul, Jumelet, Linzen,
  Mueller, Salhan, Shah, Warstadt, and Wilcox}]{babylm2026cfp}
Leshem Choshen, Ryan Cotterell, Mustafa~Omer Gul, Jaap Jumelet, Tal Linzen,
  Aaron Mueller, Suchir Salhan, Raj~Sanjay Shah, Alex Warstadt, and
  Ethan~Gotlieb Wilcox. 2026.
\newblock {BabyLM} turns 4 and goes multilingual: Call for papers for the 2026
  {BabyLM} workshop.
\newblock \emph{arXiv preprint arXiv:2602.20092}.

\bibitem[{Dehghani et~al.(2019)Dehghani, Gouws, Vinyals, Uszkoreit, and
  Kaiser}]{universaltransformer}
Mostafa Dehghani, Stephan Gouws, Oriol Vinyals, Jakob Uszkoreit, and {\L}ukasz
  Kaiser. 2019.
\newblock Universal transformers.
\newblock In \emph{International Conference on Learning Representations
  (ICLR)}.

\bibitem[{Ganescu et~al.(2025)Ganescu, Salhan, Caines, and
  Buttery}]{ganescu2025}
Bianca-Mihaela Ganescu, Suchir Salhan, Andrew Caines, and Paula Buttery. 2025.
\newblock \href {https://doi.org/10.18653/v1/2025.babylm-main.15} {Looking to
  learn: Token-wise dynamic gating for low-resource vision-language modelling}.
\newblock In \emph{Proceedings of the First BabyLM Workshop}. Association for
  Computational Linguistics.

\bibitem[{Gebru et~al.(2021)Gebru, Morgenstern, Vecchione, Vaughan, Wallach,
  Daum{\'e}~III, and Crawford}]{gebru2021datasheets}
Timnit Gebru, Jamie Morgenstern, Briana Vecchione, Jennifer~Wortman Vaughan,
  Hanna Wallach, Hal Daum{\'e}~III, and Kate Crawford. 2021.
\newblock \href {https://doi.org/10.1145/3458723} {Datasheets for datasets}.
\newblock \emph{Communications of the ACM}, 64(12):86--92.

\bibitem[{Geiping et~al.(2025)Geiping, McLeish, Jain, Kirchenbauer, Singh,
  Bartoldson, Kailkhura, Bhatele, and Goldstein}]{geiping2025depth}
Jonas Geiping, Sean McLeish, Neel Jain, John Kirchenbauer, Siddharth Singh,
  Brian~R. Bartoldson, Bhavya Kailkhura, Abhinav Bhatele, and Tom Goldstein.
  2025.
\newblock \href {https://doi.org/10.52202/085713-1380} {Scaling up test-time
  compute with latent reasoning: A recurrent depth approach}.
\newblock In \emph{Advances in Neural Information Processing Systems
  (NeurIPS)}, volume~38, pages 41340--41391. Curran Associates, Inc.

\bibitem[{Gleitman(1990)}]{gleitman1990structural}
Lila Gleitman. 1990.
\newblock The structural sources of verb meanings.
\newblock \emph{Language Acquisition}, 1(1):3--55.

\bibitem[{Goyal et~al.(2017)Goyal, Khot, Summers-Stay, Batra, and
  Parikh}]{vqav2}
Yash Goyal, Tejas Khot, Douglas Summers-Stay, Dhruv Batra, and Devi Parikh.
  2017.
\newblock Making the {V} in {VQA} matter: Elevating the role of image
  understanding in visual question answering.
\newblock In \emph{IEEE Conference on Computer Vision and Pattern Recognition
  (CVPR)}, pages 6325--6334.

\bibitem[{Hu et~al.(2024)Hu, Mueller, Ross, Williams, Linzen, Zhuang,
  Cotterell, Choshen, Warstadt, and Wilcox}]{hu2024findings}
Michael~Y. Hu, Aaron Mueller, Candace Ross, Adina Williams, Tal Linzen, Chengxu
  Zhuang, Ryan Cotterell, Leshem Choshen, Alex Warstadt, and Ethan~Gotlieb
  Wilcox. 2024.
\newblock \href {https://aclanthology.org/2024.conll-babylm.1/} {Findings of
  the second {BabyLM} challenge: Sample-efficient pretraining on
  developmentally plausible corpora}.
\newblock In \emph{Proceedings of the BabyLM Challenge at the 28th Conference
  on Computational Natural Language Learning (CoNLL)}, pages 1--21. Association
  for Computational Linguistics.

\bibitem[{Ivanova et~al.(2025)Ivanova, Sathe, Lipkin, Kumar, Radkani, Clark,
  Kauf, Hu, Pramod, Grand, Paulun, Ryskina, Aky{\"u}rek, Wilcox, Rashid,
  Choshen, Levy, Fedorenko, Tenenbaum, and Andreas}]{ewok2024}
Anna~A. Ivanova, Aalok Sathe, Benjamin Lipkin, Unnathi~U. Kumar, Setayesh
  Radkani, Thomas~H. Clark, Carina Kauf, Jennifer Hu, R.~T. Pramod, Gabriel
  Grand, Vivian~C. Paulun, Maria Ryskina, Ekin Aky{\"u}rek, Ethan~G. Wilcox,
  Nafisa Rashid, Leshem Choshen, Roger Levy, Evelina Fedorenko, Joshua
  Tenenbaum, and Jacob Andreas. 2025.
\newblock \href {https://doi.org/10.1162/tacl.a.38} {Elements of world
  knowledge ({EW}o{K}): A cognition-inspired framework for evaluating basic
  world knowledge in language models}.
\newblock \emph{Transactions of the Association for Computational Linguistics},
  13:1245--1270.

\bibitem[{Kim and Schuster(2023)}]{entitytracking}
Najoung Kim and Sebastian Schuster. 2023.
\newblock \href {https://doi.org/10.18653/v1/2023.acl-long.213} {Entity
  tracking in language models}.
\newblock In \emph{Proceedings of the 61st Annual Meeting of the Association
  for Computational Linguistics (Volume 1: Long Papers)}, pages 3835--3855,
  Toronto, Canada. Association for Computational Linguistics.

\bibitem[{Klerings et~al.(2024)Klerings, Bartelt, and
  Mueller}]{klerings2024modular}
Alina Klerings, Christian Bartelt, and Aaron Mueller. 2024.
\newblock Developmentally plausible multimodal language models are highly
  modular.
\newblock In \emph{The 2nd BabyLM Challenge at the 28th Conference on
  Computational Natural Language Learning (CoNLL)}, pages 118--139, Miami, FL,
  USA. Association for Computational Linguistics.

\bibitem[{Lakens(2017)}]{lakens2017equivalence}
Dani{\"e}l Lakens. 2017.
\newblock Equivalence tests: A practical primer for $t$ tests, correlations,
  and meta-analyses.
\newblock \emph{Social Psychological and Personality Science}, 8(4):355--362.

\bibitem[{Lin et~al.(2014)Lin, Maire, Belongie, Hays, Perona, Ramanan,
  Doll{\'a}r, and Zitnick}]{mscoco}
Tsung-Yi Lin, Michael Maire, Serge Belongie, James Hays, Pietro Perona, Deva
  Ramanan, Piotr Doll{\'a}r, and C.~Lawrence Zitnick. 2014.
\newblock {M}icrosoft {COCO}: Common objects in context.
\newblock In \emph{European Conference on Computer Vision (ECCV)}, pages
  740--755. Springer.

\bibitem[{Liu et~al.(2024)Liu, Li, Li, and Lee}]{llava15}
Haotian Liu, Chunyuan Li, Yuheng Li, and Yong~Jae Lee. 2024.
\newblock Improved baselines with visual instruction tuning.
\newblock In \emph{IEEE/CVF Conference on Computer Vision and Pattern
  Recognition (CVPR)}.

\bibitem[{Liu et~al.(2023)Liu, Li, Wu, and Lee}]{llava}
Haotian Liu, Chunyuan Li, Qingyang Wu, and Yong~Jae Lee. 2023.
\newblock Visual instruction tuning.
\newblock In \emph{Advances in Neural Information Processing Systems
  (NeurIPS)}.

\bibitem[{Loshchilov and Hutter(2017)}]{sgdr}
Ilya Loshchilov and Frank Hutter. 2017.
\newblock {SGDR}: Stochastic gradient descent with warm restarts.
\newblock In \emph{International Conference on Learning Representations
  (ICLR)}.

\bibitem[{Loshchilov and Hutter(2019)}]{adamw}
Ilya Loshchilov and Frank Hutter. 2019.
\newblock Decoupled weight decay regularization.
\newblock In \emph{International Conference on Learning Representations
  (ICLR)}.

\bibitem[{MacWhinney(2000)}]{childes2000}
Brian MacWhinney. 2000.
\newblock \emph{The {CHILDES} Project: Tools for Analyzing Talk}, 3rd edition.
\newblock Lawrence Erlbaum Associates.

\bibitem[{Misra et~al.(2023)Misra, Rayz, and Ettinger}]{comps2023}
Kanishka Misra, Julia Rayz, and Allyson Ettinger. 2023.
\newblock {COMPS}: Conceptual minimal pair sentences for testing robust
  property knowledge and its inheritance in pre-trained language models.
\newblock In \emph{Proceedings of the 17th Conference of the European Chapter
  of the Association for Computational Linguistics}. Association for
  Computational Linguistics.

\bibitem[{Oquab et~al.(2024)Oquab, Darcet, Moutakanni, Vo, Szafraniec,
  Khalidov, Fernandez, Haziza, Massa, El-Nouby, Assran, Ballas, Galuba, Howes,
  Huang, Li, Misra, Rabbat, Sharma, Synnaeve, Xu, Jegou, Mairal, Labatut,
  Joulin, and Bojanowski}]{dinov2}
Maxime Oquab, Timoth{\'e}e Darcet, Th{\'e}o Moutakanni, Huy Vo, Marc
  Szafraniec, Vasil Khalidov, Pierre Fernandez, Daniel Haziza, Francisco Massa,
  Alaaeldin El-Nouby, Mahmoud Assran, Nicolas Ballas, Wojciech Galuba, Russell
  Howes, Po-Yao Huang, Shang-Wen Li, Ishan Misra, Michael Rabbat, Vasu Sharma,
  and 7 others. 2024.
\newblock {DINOv2}: Learning robust visual features without supervision.
\newblock \emph{Transactions on Machine Learning Research (TMLR)}.

\bibitem[{Pinker(1984)}]{pinker1984language}
Steven Pinker. 1984.
\newblock \emph{Language Learnability and Language Development}.
\newblock Harvard University Press, Cambridge, MA.

\bibitem[{Pont-Tuset et~al.(2020)Pont-Tuset, Uijlings, Changpinyo, Soricut, and
  Ferrari}]{localizednarratives}
Jordi Pont-Tuset, Jasper Uijlings, Soravit Changpinyo, Radu Soricut, and
  Vittorio Ferrari. 2020.
\newblock Connecting vision and language with localized narratives.
\newblock In \emph{European Conference on Computer Vision (ECCV)}, pages
  647--664. Springer.

\bibitem[{Press and Wolf(2017)}]{weighttying}
Ofir Press and Lior Wolf. 2017.
\newblock \href {https://aclanthology.org/E17-2025/} {Using the output
  embedding to improve language models}.
\newblock In \emph{Proceedings of the 15th Conference of the {E}uropean Chapter
  of the Association for Computational Linguistics: Volume 2, Short Papers},
  pages 157--163, Valencia, Spain. Association for Computational Linguistics.

\bibitem[{Radford et~al.(2019)Radford, Wu, Child, Luan, Amodei, and
  Sutskever}]{gpt2}
Alec Radford, Jeffrey Wu, Rewon Child, David Luan, Dario Amodei, and Ilya
  Sutskever. 2019.
\newblock \href
  {https://cdn.openai.com/better-language-models/language_models_are_unsupervised_multitask_learners.pdf}
  {Language models are unsupervised multitask learners}.
\newblock OpenAI technical report.

\bibitem[{Saunshi et~al.(2025)Saunshi, Dikkala, Li, Kumar, and
  Reddi}]{saunshi2025looped}
Nikunj Saunshi, Nishanth Dikkala, Zhiyuan Li, Sanjiv Kumar, and Sashank~J.
  Reddi. 2025.
\newblock Reasoning with latent thoughts: On the power of looped transformers.
\newblock In \emph{International Conference on Learning Representations
  (ICLR)}.

\bibitem[{Schwethelm et~al.(2026)Schwethelm, Rueckert, and
  Kaissis}]{schwethelm2026isodepth}
Kristian Schwethelm, Daniel Rueckert, and Georgios Kaissis. 2026.
\newblock How much is one recurrence worth? iso-depth scaling laws for looped
  language models.
\newblock \emph{arXiv preprint arXiv:2604.21106}.

\bibitem[{Sennrich et~al.(2016)Sennrich, Haddow, and Birch}]{bpe}
Rico Sennrich, Barry Haddow, and Alexandra Birch. 2016.
\newblock \href {https://doi.org/10.18653/v1/P16-1162} {Neural machine
  translation of rare words with subword units}.
\newblock In \emph{Proceedings of the 54th Annual Meeting of the Association
  for Computational Linguistics (Volume 1: Long Papers)}, pages 1715--1725,
  Berlin, Germany. Association for Computational Linguistics.

\bibitem[{Sharma et~al.(2018)Sharma, Ding, Goodman, and Soricut}]{cc3m}
Piyush Sharma, Nan Ding, Sebastian Goodman, and Radu Soricut. 2018.
\newblock \href {https://doi.org/10.18653/v1/P18-1238} {Conceptual captions: A
  cleaned, hypernymed, image alt-text dataset for automatic image captioning}.
\newblock In \emph{Proceedings of the 56th Annual Meeting of the Association
  for Computational Linguistics (Volume 1: Long Papers)}, pages 2556--2565,
  Melbourne, Australia. Association for Computational Linguistics.

\bibitem[{Su et~al.(2024)Su, Lu, Pan, Murtadha, Wen, and Liu}]{rope}
Jianlin Su, Yu~Lu, Shengfeng Pan, Ahmed Murtadha, Bo~Wen, and Yunfeng Liu.
  2024.
\newblock {RoFormer}: Enhanced transformer with rotary position embedding.
\newblock \emph{Neurocomputing}, 568.

\bibitem[{Takmaz et~al.(2025)Takmaz, Bylinina, and
  Dotla{\v{c}}il}]{takmaz2025merging}
Ece Takmaz, Lisa Bylinina, and Jakub Dotla{\v{c}}il. 2025.
\newblock Model merging to maintain language-only performance in
  developmentally plausible multimodal models.
\newblock In \emph{Proceedings of the First BabyLM Workshop}. Association for
  Computational Linguistics.

\bibitem[{Touvron et~al.(2023)Touvron, Lavril, Izacard, Martinet, Lachaux,
  Lacroix, Rozi{\`e}re, Goyal, Hambro, Azhar, Rodriguez, Joulin, Grave, and
  Lample}]{llama}
Hugo Touvron, Thibaut Lavril, Gautier Izacard, Xavier Martinet, Marie-Anne
  Lachaux, Timoth{\'e}e Lacroix, Baptiste Rozi{\`e}re, Naman Goyal, Eric
  Hambro, Faisal Azhar, Aurelien Rodriguez, Armand Joulin, Edouard Grave, and
  Guillaume Lample. 2023.
\newblock {LLaMA}: Open and efficient foundation language models.
\newblock \emph{arXiv preprint arXiv:2302.13971}.

\bibitem[{Vaswani et~al.(2017)Vaswani, Shazeer, Parmar, Uszkoreit, Jones,
  Gomez, Kaiser, and Polosukhin}]{transformer}
Ashish Vaswani, Noam Shazeer, Niki Parmar, Jakob Uszkoreit, Llion Jones,
  Aidan~N. Gomez, {\L}ukasz Kaiser, and Illia Polosukhin. 2017.
\newblock Attention is all you need.
\newblock In \emph{Advances in Neural Information Processing Systems
  (NeurIPS)}, volume~30, pages 5998--6008.

\bibitem[{Warstadt et~al.(2023)Warstadt, Mueller, Choshen, Wilcox, Zhuang,
  Ciro, Mosquera, Paranjabe, Williams, Linzen, and
  Cotterell}]{warstadt2023findings}
Alex Warstadt, Aaron Mueller, Leshem Choshen, Ethan Wilcox, Chengxu Zhuang,
  Juan Ciro, Rafael Mosquera, Bhargavi Paranjabe, Adina Williams, Tal Linzen,
  and Ryan Cotterell. 2023.
\newblock \href {https://doi.org/10.18653/v1/2023.conll-babylm.1} {Findings of
  the {BabyLM} challenge: Sample-efficient pretraining on developmentally
  plausible corpora}.
\newblock In \emph{Proceedings of the BabyLM Challenge at the 27th Conference
  on Computational Natural Language Learning (CoNLL)}, pages 1--34, Singapore.
  Association for Computational Linguistics.

\bibitem[{Warstadt et~al.(2020)Warstadt, Parrish, Liu, Mohananey, Peng, Wang,
  and Bowman}]{blimp}
Alex Warstadt, Alicia Parrish, Haokun Liu, Anhad Mohananey, Wei Peng, Sheng-Fu
  Wang, and Samuel~R. Bowman. 2020.
\newblock {BLiMP}: The benchmark of linguistic minimal pairs for {English}.
\newblock \emph{Transactions of the Association for Computational Linguistics
  (TACL)}, 8:377--392.

\bibitem[{Xu et~al.(2026)Xu, Gao, Wang, Li, Chen, Guo, Yang, and
  Zhang}]{xu2026recursivevlm}
Ruihan Xu, Yuting Gao, Lan Wang, Jianing Li, Weihao Chen, Qingpei Guo, Ming
  Yang, and Shiliang Zhang. 2026.
\newblock Looping back to move forward: Recursive transformers for efficient
  and flexible large multimodal models.
\newblock \emph{arXiv preprint arXiv:2602.09080}.

\bibitem[{Zhang et~al.(2026)Zhang, Yan, Jiang, and Han}]{zhang2026hive}
Yiming Zhang, Qiangyu Yan, Borui Jiang, and Kai Han. 2026.
\newblock Multimodal latent reasoning via hierarchical visual cues injection.
\newblock \emph{arXiv preprint arXiv:2602.05359}.

\end{thebibliography}

\clearpage
\appendix

\section{Reproducibility}
\label{app:repro}

\begin{table}[h]
\centering
\small
\begin{tabular}{lrrr}
\toprule
Cell & $K$ & Params & Core FLOPs \\
\midrule
\texttt{std\_text}  & 1 & 97{,}241{,}856 & $1\times$ \\
\texttt{std\_mm}    & 1 & 98{,}423{,}040 & $1\times$ \\
\texttt{loop\_text} & 4 & 97{,}241{,}856 & $4\times$ \\
\texttt{loop\_mm}   & 4 & 98{,}423{,}040 & $4\times$ \\
\bottomrule
\end{tabular}
\caption{Parameter counts and relative core-stack FLOPs per token
($d_{\text{model}}{=}768$, 12 layers, 12 heads, vocab 16{,}000). Multimodal
cells differ from text cells only by the MLP connector ($+1{,}181{,}184$
parameters); \texttt{std\_mm\_id} removes it. The frozen vision encoder runs
offline and is excluded.}
\label{tab:params}
\end{table}

We save checkpoints at 28 milestones defined by cumulative training
words: every 1M words from 1M to 9M, every 10M words from 10M to 90M, and
every 100M words from 100M to 1B. Each checkpoint uses a Hugging
Face--compatible format and can be loaded with
\texttt{AutoModelForCausalLM} using \texttt{trust\_remote\_code=True}.
For each of the four submitted conditions, we publish the submitted seed's
final model and all 28 milestone checkpoints in a single Hugging Face
repository. The two additional main conditions, \texttt{std\_mm\_id} and
\texttt{loop\_mm\_prefix}, are evaluated from local checkpoints using the
same pipeline (Section~\ref{sec:eval}).

We use the official evaluation pipeline with the \texttt{causal}
backend. The original BLiMP records of the six main
conditions (Section~\ref{sec:conditions}) were produced earlier at
revision \texttt{68cdd16}; the deeper standard models were evaluated at
\texttt{3d57ddc}.
For Appendix~\ref{app:packedcontrols}, we use re-evaluated
\texttt{loop\_text} and \texttt{loop\_mm\_prefix} scores at
\texttt{3d57ddc}, matching the revision used for the loop-side
batch-matched no-image control. Tables~\ref{tab:results} and~\ref{tab:perseed} retain the
original main-condition scores.
For text-only inputs, unit tests check numerical agreement with a
single-pass standard Transformer when $K=1$ and text-input injection is
disabled. They compare output scores and input-embedding gradients.
Separate tests check the word-based learning-rate schedule in
Eq.~\ref{eq:lr}.

We train a byte-level BPE tokenizer with a vocabulary of 16,000 tokens on
the 100M-word Strict text corpus. The tokenizer is then kept fixed across
all conditions. Using the same tokenizer keeps the segmentation rules
fixed across conditions. It does not guarantee that tokenization affects
both architectures in the same way.

The assembled multimodal training corpus is available on Hugging Face,
together with a datasheet \citep{gebru2021datasheets} and provenance
records.\footnote{\url{https://huggingface.co/datasets/daichi812/babyloop-mm-corpus}}
The precomputed DINOv2 features used in the main experiments remain at
the official BabyLM distribution site.\footnote{\url{https://osf.io/ad7qg/}}

\section{Leaderboard context}
\label{app:leaderboard}
Our submitted text-only looped model scores 76.62 on BLiMP under causal
scoring on the public leaderboard. This exceeds the listed GPT-2
\citep{gpt2} Strict baseline (74.73) and Interaction baseline (72.89).
The three official GPT-BERT baseline variants score between 74.56 and
79.29 under the same causal scoring method. Our submitted model falls
within this range, below the strongest variant.
GPT-BERT \citep{gptbert} combines masked and causal language modeling
during training. These comparisons use individual submitted checkpoints
and do not control for differences in training setup. Our conclusions
rely on the controlled comparisons within our own implementation.

\section{Per-task results}
\label{app:pertask}

Table~\ref{tab:pertask} summarizes the changes in accuracy from text-only
to multimodal training and the corresponding differences between
architectures. Table~\ref{tab:pertaskfull} reports the accuracy of each
condition as the mean and standard deviation across seeds 42, 43, and 44.

Entity tracking uses the updated official evaluation, which excludes
items containing a ``nothing'' answer option.

\begin{table}[h]
\centering
\footnotesize
\setlength{\tabcolsep}{2pt}
\begin{tabular}{lccc}
\toprule
Condition & BLiMP & Suppl. & EWoK \\
\midrule
\texttt{std\_text}       & $75.21{\pm}0.68$ & $63.37{\pm}0.45$ & $53.92{\pm}0.85$ \\
\texttt{loop\_text}      & $76.16{\pm}0.45$ & $63.56{\pm}0.68$ & $54.19{\pm}0.50$ \\
\texttt{std\_mm}         & $71.37{\pm}0.29$ & $59.71{\pm}0.29$ & $53.96{\pm}0.46$ \\
\texttt{loop\_mm}        & $72.38{\pm}0.83$ & $58.61{\pm}1.53$ & $53.26{\pm}0.30$ \\
\texttt{std\_mm\_id}     & $72.48{\pm}0.69$ & $58.78{\pm}2.31$ & $53.56{\pm}0.19$ \\
\texttt{loop\_mm\_prefix}& $72.31{\pm}0.63$ & $59.82{\pm}1.35$ & $53.58{\pm}0.76$ \\
\midrule
Condition & COMPS & Entity & \\
\midrule
\texttt{std\_text}       & $56.14{\pm}0.47$ & $18.49{\pm}0.88$ & \\
\texttt{loop\_text}      & $55.87{\pm}0.17$ & $17.00{\pm}0.53$ & \\
\texttt{std\_mm}         & $55.51{\pm}0.24$ & $18.07{\pm}0.75$ & \\
\texttt{loop\_mm}        & $55.86{\pm}0.44$ & $18.85{\pm}0.57$ & \\
\texttt{std\_mm\_id}     & $55.51{\pm}0.27$ & $18.47{\pm}0.36$ & \\
\texttt{loop\_mm\_prefix}& $55.14{\pm}0.24$ & $18.45{\pm}1.74$ & \\
\bottomrule
\end{tabular}
\caption{Zero-shot accuracy (\%) on the final checkpoints, reported as
the mean and standard deviation across three seeds. All scores use the
official evaluation pipeline, and the BLiMP results match
Table~\ref{tab:results}.}
\label{tab:pertaskfull}
\end{table}

\section{Per-phenomenon decomposition}
\label{app:phenom}
\begin{figure}[h]
\centering
\includegraphics[width=\linewidth]{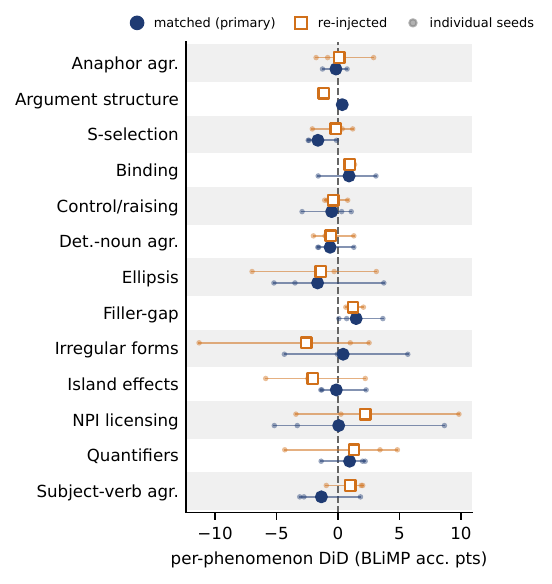}
\caption{Interactions across the 13 BLiMP linguistic categories.
Filled circles show three-seed means for the primary comparison with
matched visual input, and open squares show means for the visual-refresh
variant. Small dots show individual seed estimates; whiskers span their
minimum and maximum values.}
\label{fig:did}
\end{figure}

Figure~\ref{fig:did} reports the difference-in-differences for each of
the 13 linguistic categories in BLiMP. It shows both the primary comparison
with matched visual input and the variant with visual refresh during
looping.

The categories follow the standard BLiMP order. Mean primary interactions
range from $-1.66$ to $1.48$ BLiMP points, and their signs change across
seeds in 10 of the 13 categories. This analysis is exploratory, and we do
not adjust for multiple comparisons.

\section{Per-seed results}
\label{app:seeds}
Table~\ref{tab:perseed} reports the BLiMP score for each seed in the six
main conditions. These scores underlie the means and standard deviations
reported in Table~\ref{tab:results}.

\begin{table}[h]
\centering
\small
\setlength{\tabcolsep}{4pt}
\begin{tabular}{lcccc}
\toprule
Condition & s42 & s43 & s44 & mean$\pm$sd \\
\midrule
\texttt{std\_text}   & 75.23 & 75.88 & 74.53 & $75.21{\pm}0.68$ \\
\texttt{loop\_text}  & 76.61 & 75.71 & 76.17 & $76.16{\pm}0.45$ \\
\texttt{std\_mm}     & 71.62 & 71.05 & 71.44 & $71.37{\pm}0.29$ \\
\texttt{std\_mm\_id} & 71.68 & 72.93 & 72.82 & $72.48{\pm}0.69$ \\
\texttt{loop\_mm}    & 71.56 & 73.21 & 72.36 & $72.38{\pm}0.83$ \\
\texttt{loop\_mm\_prefix} & 72.34 & 71.66 & 72.92 & $72.31{\pm}0.63$ \\
\bottomrule
\end{tabular}
\caption{BLiMP accuracy (\%) on the final checkpoints for seeds 42, 43,
and 44. The looped model scores below the standard model for text-only
seed 43 and for multimodal seed 42 with visual refresh. With matched
visual input, the looped model scores higher for all three seeds.}
\label{tab:perseed}
\end{table}

\section{Learning trajectories}
\label{app:traj}
\begin{figure}[t]
\centering
\includegraphics[width=\linewidth]{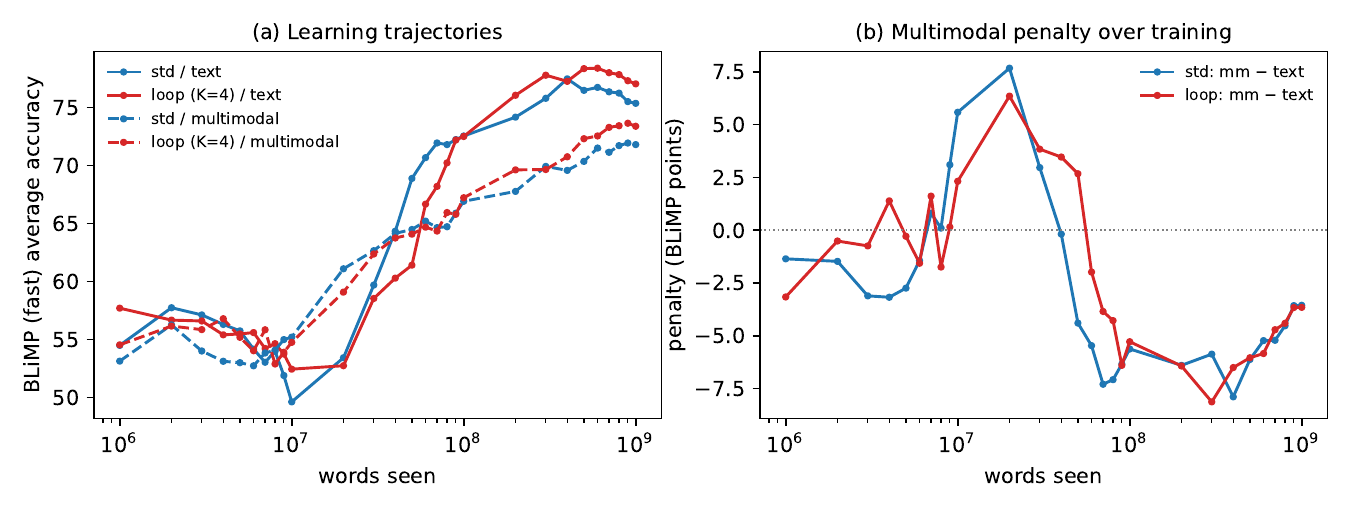}
\caption{(a) BLiMP accuracy on the fast-evaluation subset at 28
checkpoints, using one submitted seed per condition. (b) Multimodal minus
text-only accuracy within each architecture. Both panels plot cumulative
training words on a logarithmic horizontal axis.}
\label{fig:traj}
\end{figure}

Figure~\ref{fig:traj} tracks BLiMP accuracy across 28 checkpoints for the
four submitted models. The curves use the fast-evaluation subset and one
seed per condition, so they describe individual training runs. The
horizontal axis shows cumulative training words.
Both multimodal models briefly outperform their text-only counterparts
early in training. The text-only models later overtake them, and the gap
fluctuates rather than narrowing steadily. By the end of training,
multimodal accuracy is about 3.6 BLiMP points lower in both architectures.
The early advantage could reflect differences in the language data, the
visual input, or both. The trajectories alone do not separate these
contributions.

\begin{figure}[t]
\centering
\includegraphics[width=\linewidth]{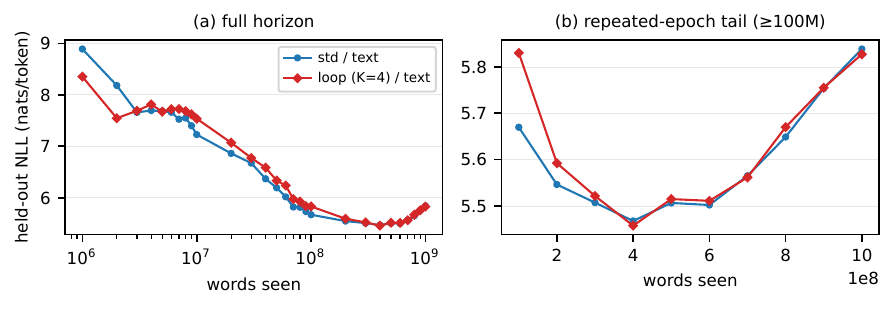}
\caption{Held-out caption loss (nats per token) for the standard and
looped text-only models, using 20,000 captions and seed 42. Cumulative
training words are plotted on a log scale for all 28 checkpoints in (a),
and on a linear scale from 100M onward in (b).}
\label{fig:heldout}
\end{figure}

Figure~\ref{fig:heldout} tracks held-out caption loss for the standard
and looped text-only models, using seed 42. We evaluate all 28 checkpoints
on the same 20,000 captions selected after the training-caption cutoff.
Loss is measured as negative log-likelihood in nats per token.
Both models reach their lowest held-out loss at 400M cumulative training
words. By the end of training, loss has increased by about 0.37 nats per
token in both models. This pattern is consistent with overfitting during
later training.
At the final checkpoint, held-out loss is 5.84 for the standard model and
5.83 for the looped model. Thus, the looped model's higher BLiMP accuracy
is accompanied by little difference in held-out caption loss.

\section{Shuffled-features diagnostic}
\label{app:shufflediag}
We evaluate the final checkpoints of the four main multimodal conditions,
using three seeds per condition. All models are evaluated on the same
fixed sample of 50,000 training captions. We change the visual features
only at evaluation time, without retraining the models.
Each caption is evaluated with its correct image features, features
shuffled across the sampled captions, and features replaced by zeros.
The shuffled result is averaged over three fixed permutations. Caption
text, batching, and numerical precision are held constant across these
conditions.

For each caption, we calculate the average token-level negative
log-likelihood using the correct preceding tokens. Visual prefix
positions are excluded from the loss. The shuffled-minus-correct
difference is calculated separately for each caption.
Because the captions come from the training data, this diagnostic
measures visual reliance on familiar examples. It does not establish
generalization to unseen image--caption pairs.

\begin{table}[h]
\centering
\small
\setlength{\tabcolsep}{4pt}
\begin{tabular}{lcccc}
\toprule
Cell & correct & shuf. & zero & $\Delta_{\text{shuf}}$ \\
\midrule
\texttt{std\_mm}          & 1.87 & 3.33 & 3.78 & $1.453{\pm}0.005$ \\
\texttt{std\_mm\_id}      & 1.87 & 3.32 & 3.31 & $1.451{\pm}0.006$ \\
\texttt{loop\_mm}         & 1.84 & 3.35 & 3.74 & $1.516{\pm}0.002$ \\
\texttt{loop\_mm\_prefix} & 1.85 & 3.34 & 3.78 & $1.485{\pm}0.009$ \\
\bottomrule
\end{tabular}
\caption{Caption negative log-likelihood (nats per token) under correct,
shuffled, and zeroed image features, averaged equally across captions and
then across three training seeds. $\Delta_{\mathrm{shuf}}$ reports the
shuffled-minus-correct difference as the mean and standard deviation
across training seeds.}
\label{tab:shufflediag}
\end{table}

For every model and seed, at least 98.6\% of captions have higher loss
with shuffled features, averaged over the three permutations. This shows
that correct image--caption pairing helps prediction on these training
examples.
With MLP connectors, zeroed features produce higher loss than shuffled
features. With the identity connector, the losses are similar. Zeroing
replaces the usual range of image features with a constant input, so the
loss change need not reflect only the removal of visual information.

\section{Visual question answering}
\label{app:vqa}
We use the 2024 BabyLM \texttt{vqa\_filtered} task, based on VQAv2
questions \citep{vqav2} and MS-COCO images \citep{mscoco}. Each of the
25,230 questions has eight candidate answers. Random choice therefore
gives 12.5\% accuracy.
Given an image and a question, the model scores each candidate answer by
summing its token log-probabilities. We select the candidate with the
highest score, without length normalization.

For the main conditions, we extract one global feature vector per VQA
image using frozen DINOv2 ViT-B/14. These features have the same input
format as the training features, but are extracted by us rather than
supplied by the organizers. We report the resulting scores as
diagnostics, not official leaderboard results.
Each model is evaluated with correct, shuffled, and zeroed image
features. The shuffled condition uses one fixed permutation of the
feature rows. Questions, candidate answers, and item order are kept
fixed across conditions and models.
We use an exact McNemar test to compare correct and shuffled features
within each model and seed. The test uses questions answered correctly
under one feature condition but incorrectly under the other.

\begin{table}[h]
\centering
\scriptsize
\setlength{\tabcolsep}{2.5pt}
\begin{tabular}{lcccc}
\toprule
Cell & correct & shuf. & zero & $\Delta_{\text{cor}-\text{shuf}}$ \\
\midrule
\texttt{std\_mm}          & $47.4{\pm}1.1$ & $42.9{\pm}0.6$ & $44.2{\pm}2.3$ & $+4.6{\pm}1.0$ \\
\texttt{std\_mm\_id}      & $45.1{\pm}3.6$ & $41.8{\pm}3.1$ & $42.6{\pm}3.6$ & $+3.3{\pm}1.2$ \\
\texttt{loop\_mm}         & $45.1{\pm}4.8$ & $41.0{\pm}5.3$ & $35.1{\pm}14.1$ & $+4.1{\pm}1.0$ \\
\texttt{loop\_mm\_prefix} & $41.2{\pm}4.8$ & $37.3{\pm}4.8$ & $40.2{\pm}3.5$ & $+3.8{\pm}0.6$ \\
\midrule
\texttt{std2x\_mm}        & $38.6{\pm}13.3$ & $34.9{\pm}12.4$ & $33.0{\pm}12.9$ & $+3.7{\pm}1.0$ \\
\texttt{std4x\_mm}        & $39.1{\pm}6.7$ & $35.7{\pm}5.4$ & $30.6{\pm}6.5$ & $+3.4{\pm}1.4$ \\
\midrule
\texttt{subset}$^\ddagger$   & $44.9$ & $41.8$ & $40.4$ & $+3.1$ \\
\texttt{rich}$^\ddagger$     & $45.1$ & $39.0$ & $42.2$ & $+6.1$ \\
\texttt{hive}$^\ddagger$     & $41.6$ & $35.4$ & $37.6$ & $+6.2$ \\
\texttt{hive\_k8}$^\ddagger$ & $27.1$ & $20.4$ & $24.7$ & $+6.7$ \\
\bottomrule
\end{tabular}
\caption{VQA accuracy (\%) with correct, shuffled, and zeroed image
features. $\Delta_{\mathrm{cor}-\mathrm{shuf}}$ is the correct-minus-shuffled
difference in percentage points. Values are mean $\pm$ standard deviation
across three training seeds, except for the single-seed exploratory models
($\ddagger$) trained on a reduced image subset.}
\label{tab:vqa}
\end{table}

Correct image features yield higher VQA accuracy than shuffled features
in all twelve model--seed combinations. The gains range from 2.4 to 5.6
percentage points, with $p<10^{-25}$ in every McNemar test.
With zeroed features, \texttt{loop\_mm} scores 18.9\% for seed 43,
compared with 42.7\% and 43.8\% for seeds 42 and 44. This large variation
shows that responses to zeroed features can differ substantially across
training seeds. We therefore interpret this condition cautiously when
assessing visual reliance.

\paragraph{Deeper standard models.}
The deeper standard models are trained on the same official visual
features and evaluated with the same VQA protocol as the main models.
With correct image features, the 24-layer model scores 38.6\% and the
48-layer model scores 39.1\%. Both means are below the 12-layer standard
model's 47.4\%.
Variation across seeds is substantial, with standard deviations of 13.3
percentage points for the 24-layer model and 6.7 for the 48-layer model.
These runs provide no clear evidence of a VQA benefit from increasing
standard-model depth.

\paragraph{Exploratory visual inputs.}
In Table~\ref{tab:vqa}, the common \texttt{loop\_mm\_} prefix is omitted
from the names of the four exploratory models marked $\ddagger$.
Each exploratory model is evaluated with the visual feature format used
during its training: a single global vector, patch-level vectors, or
vectors from multiple encoder layers. These models were trained on a
reduced image subset using self-extracted features
(Appendix~\ref{app:exploratory}). Comparisons with the main models
therefore combine changes in training data and visual representation.
The single-vector reference model scores 44.9\%, with a 3.1-point
advantage for correct over shuffled features. The patch-feature model
scores 45.1\%, with a 6.1-point advantage. Thus, the larger effect of
shuffling is accompanied by similar accuracy with correct images.
Hierarchical injection yields 41.6\% accuracy and a 6.2-point
correct-minus-shuffled difference. The $K=8$ training recipe yields
27.1\% accuracy and a 6.7-point difference. Although the latter still
benefits from correct image pairing, its overall VQA accuracy is
substantially lower.
These exploratory results use one training seed per condition.
Additional runs are needed to determine whether the observed differences
persist across seeds.

\section{Exploratory: richer, staged visual signals}
\label{app:exploratory}
We explore whether richer visual features and staged visual input improve
BLiMP performance in looped models. We compare four conditions, moving
from a single global image vector to training setups inspired by HIVE
\citep{zhang2026hive}.

\paragraph{Retrieving images for richer features.}
We match the released captions to their entries in Localized Narratives
and Conceptual Captions, then retrieve the corresponding images. We use
the frozen DINOv2 ViT-B/14 encoder to extract visual features from these
images.
As a spot check, we compare self-extracted and official features for 30
retrieved Localized Narratives images. All 30 pairs have cosine
similarity of at least 0.9995. The highest similarity to an unrelated
official feature vector is 0.42.

Not all captions could be matched to retrievable images. The resulting
subset contains 761,334 Localized Narratives pairs and 1,191,243
Conceptual Captions pairs. This gives 38.2M caption words, compared with
50M in the main experiments.
Localized Narratives accounts for about 71\% of the caption words in
this subset, compared with about 55\% in the main experiments. All four
exploratory conditions use the same reduced subset, with identical
caption-to-feature row mappings.
We train each condition until cumulative exposure to text and caption
words reaches 1B words. Text and caption batches are sampled with equal
probability, which does not imply equal word exposure because caption
batches contain fewer words. These experiments are analysis-only and
were not submitted to the competition.

\paragraph{Experimental conditions.}
All four conditions use MLP connectors and a peak learning rate of
$6\times10^{-4}$. They use the same word-based learning-rate schedule and
batch settings as the main experiments, with one training seed per
condition. The peak learning rate differs from the $1.5\times10^{-3}$
used for the main looped models.
\begin{itemize}[leftmargin=*, itemsep=2pt, topsep=4pt, parsep=0pt, partopsep=0pt]
  \item \texttt{loop\_mm\_subset} uses a single global feature vector from
        the final encoder layer, supplied once as a visual prefix, with
        four loop iterations. It provides a reference for the other
        exploratory conditions on the same reduced dataset.
  \item \texttt{loop\_mm\_rich} replaces the single global vector with 16
        patch-level feature vectors from the final encoder layer. These
        are supplied once as visual prefix tokens, while the model still
        uses four loop iterations.
  \item \texttt{loop\_mm\_hive} uses 16 feature vectors from each of
        encoder blocks 3, 6, 9, and 12. Each block has its own MLP
        connector. Block-3 features form the initial visual prefix, while
        features from blocks 6, 9, and 12 are added before the second,
        third, and fourth loop iterations.
        The separate connectors add approximately 3.5M parameters
        relative to \texttt{loop\_mm\_rich}. This comparison therefore
        changes the source of visual features, their injection schedule,
        and the number of connector parameters together.
  \item \texttt{loop\_mm\_hive\_k8} uses eight loop iterations and repeats
        the sequence of visual features from encoder blocks 3, 6, 9,
        and 12. Gradients are propagated only through the final four
        iterations \citep{zhang2026hive}. This comparison combines changes
        in loop count, feature scheduling, and gradient propagation.
\end{itemize}

\paragraph{BLiMP results.}
The single-vector reference scores 70.60 on BLiMP. Replacing it with
patch-level features gives 70.47, while hierarchical injection gives
70.87. These single-seed results do not establish a benefit from richer
or staged visual input.
The $K=8$ training recipe scores 55.36 on BLiMP, substantially below the
other three conditions. Because this recipe changes loop count, feature
cycling, and gradient propagation together, the decrease cannot be
attributed to any one of these changes.
The $K=8$ condition also inherits the learning rate used for the $K=4$
exploratory models. Further tuning would be needed to determine whether
its low BLiMP score persists under other learning rates.

\begin{table}[h]
\centering
\footnotesize
\setlength{\tabcolsep}{3pt}
\begin{tabular}{lcc}
\toprule
Cell & (adds) & BLiMP \\
\midrule
\texttt{loop\_mm\_subset}   & anchor      & 70.60 \\
\texttt{loop\_mm\_rich}     & $+$richness & 70.47 \\
\texttt{loop\_mm\_hive}     & $+$staging  & 70.87 \\
\texttt{loop\_mm\_hive\_k8} & $+$recipe   & 55.36 \\
\bottomrule
\end{tabular}
\caption{BLiMP accuracy (\%) on the full evaluation set for the four
exploratory conditions, each trained with seed 42. All conditions use
the same reduced image subset and a peak learning rate of
$6\times10^{-4}$. VQA results for these models are reported in
Appendix~\ref{app:vqa}, Table~\ref{tab:vqa}.}
\label{tab:exploratory}
\end{table}

\paragraph{Zeroing features from individual encoder blocks.}
To probe dependence on features from each encoder block, we zero one
block's features at evaluation time while leaving the others unchanged.
We apply this diagnostic to \texttt{loop\_mm\_hive} and
\texttt{loop\_mm\_hive\_k8}, both trained with seed 42. Each comparison
uses the same fixed subset of 6,000 VQA questions.

For \texttt{loop\_mm\_hive}, zeroing block-12 features lowers accuracy
from 40.6\% to 35.5\%. The correct-minus-shuffled difference also falls
from 6.2 to 1.5 points. Zeroing block 3 instead raises accuracy by 3.7
points, while zeroing blocks 6 or 9 changes it by about 0.1 point.
For \texttt{loop\_mm\_hive\_k8}, zeroing block-12 features lowers accuracy
from 27.1\% to 26.1\%. The correct-minus-shuffled difference falls from
6.9 to 0.6 points. Zeroing block 9 instead raises accuracy by 3.0 points.

In both models, zeroing block-12 features produces the largest reduction
in the correct-minus-shuffled difference. This suggests that these
features play an important role in the observed benefit from correct
image pairing. These single-seed, evaluation-time probes do not establish
that lower-layer features are unnecessary during training.

\section{K sweep}
\label{app:ksweep}
We compare text-only models at $K=1,2,4,$ and $8$. The $K=1$ standard
model and $K=4$ looped model are the main experimental conditions, each
trained with three seeds. The additional $K=2$ and $K=8$ conditions use
one seed each.
Text-input injection is enabled for $K=2,4,$ and $8$, and disabled for
the $K=1$ standard model. Comparisons with the standard model therefore
change both the number of iterations and text-input injection.

At the shared peak learning rate of $1.5\times10^{-3}$, the $K=8$ run
uses micro-batches of 32 sequences and 16 gradient accumulation steps.
This preserves the 512 sequences per optimizer update used by the other
conditions, which use micro-batches of 64 and eight accumulation steps.
All looped models in this text-only comparison propagate gradients
through every iteration. They use no visual input or gradient
truncation, unlike the $K=8$ multimodal recipe in
Appendix~\ref{app:exploratory}.

\paragraph{Learning-rate probes for $K=8$.}
Before starting the 1B-word runs at $K=8$, we compare four peak learning
rates in probes trained for 100M cumulative words. The candidates are
$3\times10^{-4}$, $6\times10^{-4}$, $1\times10^{-3}$, and
$1.5\times10^{-3}$. We select the rate with the lowest final training
loss among numerically stable runs.
Final training loss decreases from 3.03 to 2.86 as the learning rate
increases across the tested range. The lowest loss is obtained at
$1.5\times10^{-3}$, which is also the rate used for the main looped
model. We train the 1B-word runs at both this rate and the next-best
candidate, $1\times10^{-3}$.

\begin{table}[h]
\centering
\footnotesize
\setlength{\tabcolsep}{3pt}
\begin{tabular}{lccc@{\hspace{1em}}cc}
\toprule
\multicolumn{4}{c}{1B-word runs} & \multicolumn{2}{c}{$K{=}8$ probes} \\
\cmidrule(r{1em}){1-4} \cmidrule{5-6}
$K$ & peak LR & seeds & BLiMP & LR & loss \\
\midrule
1 & $1.5\times10^{-3}$ & 3 & $75.21{\pm}0.68$ & $3\times10^{-4}$   & $3.03$ \\
2 & $1.5\times10^{-3}$ & 1 & $75.28$ & $6\times10^{-4}$   & $2.96$ \\
4 & $1.5\times10^{-3}$ & 3 & $76.16{\pm}0.45$ & $1\times10^{-3}$   & $2.93$ \\
8 & $1.5\times10^{-3}$ & 1 & $76.35$ & $1.5\times10^{-3}$ & $2.86$ \\
8 & $1\times10^{-3}$   & 1 & $75.42$ &                    &        \\
\bottomrule
\end{tabular}
\caption{Left: BLiMP accuracy on the full evaluation set after 1B
cumulative training words. The $K=1$ and $K=4$ rows report means and
standard deviations across three seeds; the other rows report individual
runs with seed 42. Right: final training loss from the four $K=8$
learning-rate probes, each trained for 100M cumulative words.}
\label{tab:ksweep}
\end{table}

\paragraph{Results and scope.}
At the shared learning rate, $K=2$ scores 75.28 on BLiMP and $K=8$
scores 76.35. The main $K=4$ condition has a three-seed mean of 76.16
with a standard deviation of 0.45. At the alternative learning rate of
$1\times10^{-3}$, $K=8$ scores 75.42.
These results provide no clear evidence that eight iterations improve
on four under the tested settings. However, with only one seed for each
additional condition, they do not establish an optimal loop count.
The lowest probe loss occurs at the highest learning rate tested. This
does not establish that the optimal learning rate is independent of
loop count.

\section{Modality-order curriculum}
\label{app:curriculum}
We explore training schedules that gradually shift between image--caption
batches and text-only batches. The probability of selecting a text-only
batch, $p_{\text{text}}(w)$, changes with cumulative training words $w$.
In \texttt{v2l}, $p_{\text{text}}$ increases linearly from 0 at the start
of training to 1 at 1B cumulative words. In \texttt{l2v}, it decreases
linearly from 1 to 0 over the same horizon. The flat control keeps
$p_{\text{text}}=0.5$ throughout training.

Text-only and caption batches contain different numbers of words, so a
symmetric schedule of batch-selection probabilities does not ensure equal
word exposure to the two sources. These schedules should not be
interpreted as holding cumulative word exposure to each source fixed.
All curriculum conditions use the \texttt{loop\_mm\_prefix} architecture:
four iterations, an MLP connector, and one visual prefix token supplied
at the start. They use the same 50M-word text corpus, 50M-word caption
corpus, and official visual features as the flat control. We train
\texttt{v2l} and \texttt{l2v} with three seeds each, and
\texttt{v2l\_short} with one seed.

\paragraph{Reverse schedule.}
The reverse schedule lets us compare training that ends with text-only
batches against training that ends with image--caption batches. This
comparison alone does not separate effects of early training from
effects of the data seen near the end.

\paragraph{Caption-length schedule.}
\texttt{v2l\_short} adds a caption-length schedule to \texttt{v2l}.
Training begins with the shortest 10\% of captions, ranked by word count.
The available fraction increases linearly with cumulative training words,
reaching the full caption set at 1B words.
The schedule changes which captions are available, without shortening
the captions themselves. Shorter captions are not necessarily
syntactically simpler, so this is a length-based sampling scheme rather
than a direct test of syntactic complexity.

\paragraph{Predictions.}
We expected \texttt{v2l} to improve BLiMP accuracy at the cost of lower
VQA accuracy. We also expected prioritizing shorter captions in
\texttt{v2l\_short} to yield an additional BLiMP gain.

\begin{table}[t]
\centering
\scriptsize
\setlength{\tabcolsep}{2.5pt}
\begin{tabular}{llccc}
\toprule
Cell & $p_{\text{text}}$ & BLiMP & VQA & $\Delta_{\text{cor}-\text{shuf}}$ \\
\midrule
flat  & $0.5$ & $72.31{\pm}0.63$ & $41.0{\pm}4.9$ & $+3.7{\pm}0.6$ \\
\texttt{v2l}        & $0 \to 1$ & $72.70{\pm}0.47$ & $46.4{\pm}5.4$ & $+6.0{\pm}0.8$ \\
\texttt{l2v}        & $1 \to 0$ & $66.62{\pm}1.13$ & $35.5{\pm}5.6$ & $+2.8{\pm}1.9$ \\
\texttt{v2l\_short} & $0 \to 1$, len. & $70.66$ & $23.7^{\S}$ & $+3.2^{\S}$ \\
\bottomrule
\end{tabular}
\caption{BLiMP accuracy on the full evaluation set and VQA accuracy
with correct image features (\%). Scores with $\pm$ are means and
standard deviations across three seeds; scores without $\pm$ use seed 42.
$p_{\text{text}}$ is the probability of selecting a text-only batch, and
``len.''\ denotes the caption-length schedule. The VQA columns for the
three-seed conditions come from a re-evaluation of the final checkpoints
in fp32; Table~\ref{tab:vqa} reports the original bf16 evaluation of the
flat model. $^{\S}$Single-seed value from the original bf16 evaluation.
$\Delta_{\mathrm{cor}-\mathrm{shuf}}$ is the correct-minus-shuffled VQA
difference in percentage points.}
\label{tab:curriculum}
\end{table}

\paragraph{Results and scope.}
The mean BLiMP score is 72.70 for \texttt{v2l}, compared with 72.31
for the flat control. The mean for \texttt{l2v} is 66.62, about 5.7
points below the flat control. The gap between the two curriculum
directions is about 6.1 points.
All three \texttt{l2v} runs score below the corresponding flat-control
runs. The \texttt{v2l} difference from the flat control is positive for
two seeds and negative for one, so its small positive mean should be
interpreted cautiously.
\texttt{v2l\_short} scores 70.66 on BLiMP, compared with 73.20 for
\texttt{v2l} using the same seed, 42. This single-seed result does not
support the predicted benefit of prioritizing shorter captions.

Across three seeds, VQA accuracy with correct image features is 46.4\%
for \texttt{v2l}, 41.0\% for the flat control, and 35.5\% for
\texttt{l2v}. The \texttt{v2l} difference from the flat control is
positive for two seeds and negative for one. Its mean paired difference
is $+5.4$ percentage points, with a 95\% confidence interval of
$[-19.4, 30.1]$. The predicted drop in VQA accuracy under \texttt{v2l}
is not observed in the three-seed mean, but the higher mean does not establish a reliable
improvement across seeds. These scores come from a re-evaluation of the
final checkpoints in fp32 with the same protocol for all nine models; the
single-seed \texttt{v2l\_short} model scores 23.7\% under the original
evaluation.
The results show an asymmetry between the two training schedules, but do
not identify its cause. In particular, they do not rule out effects of
recent training data or differences in cumulative exposure to each source.

\section{No-image controls for the looped model}
\label{app:packedcontrols}
\paragraph{Batch-matched control.}
We train the looped model on the language data of
\texttt{loop\_mm\_prefix} with the same batches, word schedule, and
initialization, omitting only the visual prefix
(\texttt{loop\_mm\_no\_image\_matched}; Section~\ref{sec:rq3}).
Table~\ref{tab:loopcontrol} reports the per-seed
BLiMP scores and the paired contrasts, with unadjusted 95\% confidence
intervals from a $t$ distribution with df$=2$. All three loop-side
conditions used in these contrasts were evaluated at pipeline revision
\texttt{3d57ddc}.
We re-evaluate the text-only and multimodal references at that revision
and use the existing matched-control records from the same revision
(Appendix~\ref{app:repro}).\footnote{All contrasts in
Table~\ref{tab:loopcontrol} use the common-revision per-seed run records.
The re-evaluated reference scores differ from the historical scores in
Table~\ref{tab:perseed} by at most 0.01 percentage points.}

\begin{table}[h]
\centering
\scriptsize
\setlength{\tabcolsep}{2pt}
\begin{tabular}{lrrrrl}
\toprule
 & s42 & s43 & s44 & mean$\pm$sd & 95\% CI \\
\midrule
control (looped) & 70.12 & 71.38 & 69.88 & $70.46{\pm}0.81$ & \\
\midrule
\multicolumn{6}{l}{\emph{Visual-prefix effect (multimodal $-$ matched control)}} \\
looped   & $+2.23$ & $+0.28$ & $+3.04$ & $+1.85{\pm}1.42$ & $[-1.67,5.37]$ \\
\midrule
\multicolumn{6}{l}{\emph{Language-data and batching change (text-only $-$ matched control)}} \\
looped   & $6.50$ & $4.33$ & $6.29$ & $5.71{\pm}1.20$ & $[2.73,8.68]$ \\
\bottomrule
\end{tabular}
\caption{Loop-side batch-matched no-image control, BLiMP accuracy (\%)
and paired contrasts in percentage points. The comparisons use
\texttt{loop\_mm\_prefix}, \texttt{loop\_mm\_no\_image\_matched}, and
\texttt{loop\_text}, with three seeds per condition.}
\label{tab:loopcontrol}
\end{table}

\paragraph{Packed controls.}
We also train a looped text-only model on the multimodal conditions'
language data, packing text and captions into fixed-length blocks.
This control, \texttt{loop\_text\_mmdata}, scores $71.31\pm0.20$ on BLiMP
across three seeds.\footnote{The per-seed scores are 71.54, 71.20, and
71.18 for seeds 42, 43, and 44.}
The corresponding standard-model packed control scores $70.56\pm0.19$
(Section~\ref{sec:rq3}).\footnote{The standard-model scores are 70.34,
70.67, and 70.68 for seeds 42, 43, and 44. The first two records lack
an evaluation-revision identifier; seed 44 and all three looped packed
controls record \texttt{3d57ddc}. We treat the packed comparisons as
supporting evidence, separate from the common-revision decomposition.}
Both controls score below their counterparts
trained on the original text-only corpus (Table~\ref{tab:results}).
These results are consistent with changes in language data contributing
to the BLiMP drop in both architectures.
Under the fixed word budget, the standard packed control uses 6,221
optimizer updates, compared with 6,827 for the original text-only model.
Thus the data change is not isolated from its consequences for the
number of updates.

The packed controls use a different batching procedure from multimodal
training. Comparisons with the multimodal models therefore combine
changes in batching and visual input, rather than isolating the effect
of vision. The looped packed control does not replace the batch-matched
control described in Limitations.

\end{document}